%% file: Pix2Surf_ECCV2020.tex
\documentclass[runningheads]{llncs}
\usepackage{graphicx}
\usepackage{comment}
\usepackage{amsmath,amssymb} 
\usepackage{color}
\usepackage{hyperref}
\usepackage{siunitx}
\usepackage{microtype}
\usepackage{relsize}
\input{macros.tex}

\usepackage[width=122mm,left=12mm,paperwidth=146mm,height=193mm,top=12mm,paperheight=217mm]{geometry}

\begin{document}
\pagestyle{headings}
\mainmatter
\def\ECCVSubNumber{2941}  

\title{Pix2Surf: Learning Parametric 3D Surface Models of Objects from Images}

\titlerunning{Pix2Surf}
\authorrunning{J. Lei et al.}
%
\author{Jiahui Lei\textsuperscript{1} \quad Srinath Sridhar\textsuperscript{2} \quad Paul Guerrero\textsuperscript{3} \quad Minhyuk Sung\textsuperscript{3}\\ \quad Niloy Mitra\textsuperscript{4,3} \quad Leonidas J.~Guibas\textsuperscript{2}}
\institute{
\textsuperscript{1}Zhejiang University \quad  \textsuperscript{2}Stanford University \quad  \textsuperscript{3}Adobe Research \quad \\ \textsuperscript{4}University College London\\ \ \\
\faGlobe~\href{https://geometry.stanford.edu/projects/pix2surf}{https://geometry.stanford.edu/projects/pix2surf}
}

\maketitle

\input{content/text/00_abstract}
\input{content/text/01_introduction}
\input{content/text/02_relwork}
\input{content/text/03_overview}
\input{content/text/04_method}
\input{content/text/04_1_multi}
\input{content/text/05_experiments_new}
\input{content/text/06_conclusion}

{\small
\bibliographystyle{splncs04}
\bibliography{references}
}

\clearpage

\renewcommand{\thesection}{S}
\setcounter{table}{0}
\renewcommand{\thetable}{S\arabic{table}}
\setcounter{figure}{0}
\renewcommand{\thefigure}{S\arabic{figure}}

\newif\ifpaper
\papertrue

\section*{Supplementary Material}
\input{content/supp/text/10_supplementary.tex}

\end{document}

%% file: macros.tex
\usepackage{amsmath}
\usepackage{amssymb}
\usepackage{booktabs}
\usepackage{capt-of} 
\usepackage{cuted} 
\usepackage{etoolbox} 
\usepackage{float}
\usepackage{fontawesome}
\usepackage{graphicx}
\usepackage{makecell}
\usepackage{marvosym}
\usepackage{multirow}
\usepackage{pifont}
\usepackage{tabularx}
\usepackage{wrapfig}
\usepackage[dvipsnames]{xcolor}

\usepackage{catchfile}
\usepackage{longtable}

\newcommand{\xmark}{\ding{55}}%

\newcommand{\ie}{i.e.,}
\newcommand{\eg}{e.g.,}

\newcommand{\parahead}[1]{\noindent\textbf{#1}:\ }

\newenvironment{packed_itemize}
{\begin{itemize}
    \setlength{\itemsep}{1pt}
    \setlength{\parskip}{0pt}
    \setlength{\parsep}{0pt}
}{\end{itemize}}

\newcommand{\filluptopage}[1]{%
  \clearpage
  \loop\ifnum\value{page}<#1\relax
    \null\clearpage
  \repeat
  \loop\ifnum\value{page}=#1\relax
    \null\clearpage
  \repeat
}

\makeatletter
\def\blfootnote{\xdef\@thefnmark{}\@footnotetext}
\makeatother

\newcommand{\jiahui}[1]{{#1}}

\newcommand{\minhyuk}[1]{{#1}}

%% file: content/text/00_abstract.tex
\begin{abstract}
We investigate the problem of learning to generate 3D parametric surface representations for novel object instances, as seen from one or more views. Previous work on learning shape reconstruction from multiple views uses discrete representations such as point clouds or voxels, while continuous surface generation approaches lack multi-view consistency. We address these issues by designing neural networks capable of generating \emph{high-quality parametric 3D surfaces} which are also \emph{consistent} between views. Furthermore, the generated 3D surfaces preserve accurate image pixel to 3D surface point \emph{correspondences}, allowing us to lift texture information to reconstruct shapes with rich geometry \emph{and} appearance. Our method is supervised and trained on a public dataset of shapes from common object categories. Quantitative results indicate that our method significantly outperforms previous work, while qualitative results demonstrate the high quality of our reconstructions.
\keywords{3D reconstruction, multi-view, single-view, parametrization}
\end{abstract}

%% file: content/text/01_introduction.tex
\section{Introduction}
Reconstructing the 3D shape of an object from one or more views is an important problem with applications in 3D scene understanding, robotic navigation or manipulation, and content creation.
Even with multi-view images, the problem can be challenging when camera baselines are large, or when lighting and occlusions are inconsistent across the views.
Recent developments in supervised deep learning have demonstrated the potential to overcome these challenges. 

Ideally, a multi-view surface reconstruction algorithm should have the following desirable \textbf{\underline{3C}} properties: \jiahui{surface \textbf{\underline{c}}ontinuity, multi-view \textbf{\underline{c}}onsistency and \mbox{2D-3D} \textbf{\underline{c}}orrespondence}.
First, it should be able to reconstruct high-quality shapes that can be readily used in downstream applications.
While much progress has been made in learning shape representations such as point clouds~\cite{Fan:2017,lin2018learning,Srinath:2019,insafutdinov18pointclouds}, volumetric grids~\cite{choy20163d,tatarchenko2019single,drcTulsiani17}, and meshes~\cite{Wang:2018:pixel2mesh,Wen:2019:pixel2mesh++}, their geometric quality is limited by the discrete nature of the underlying representation.
Therefore, representations such as implicit functions~\cite{Park:2019,saito2019pifu,Chen:2019:implicit}, and $UV$ surface parametrizations~\cite{Groueix:2018,Deprelle:2019} are preferable, since they can represent a \textbf{continuous surface} at arbitrary resolution.
Second, the algorithm should be able to reconstruct objects from a sparse set of views while ensuring that the combined shape is \textbf{consistent} across the views.
Recent approaches exploit geometric constraints to solve this problem but require additional supervision through knowledge of the exact camera geometry~\cite{ChenPMVSNet2019ICCV}.
Finally, the algorithm should provide accurate \textbf{correspondences} between 2D pixels and points on the 3D shape, so as to accurately transport object properties (\mbox{\eg~texture}) directly from 2D and support aggregation across views.
While some extant methods satisfy a subset of these properties, we currently lack any method that has all of them.

\begin{figure}[t]
    \centering
    \includegraphics[width=\columnwidth]{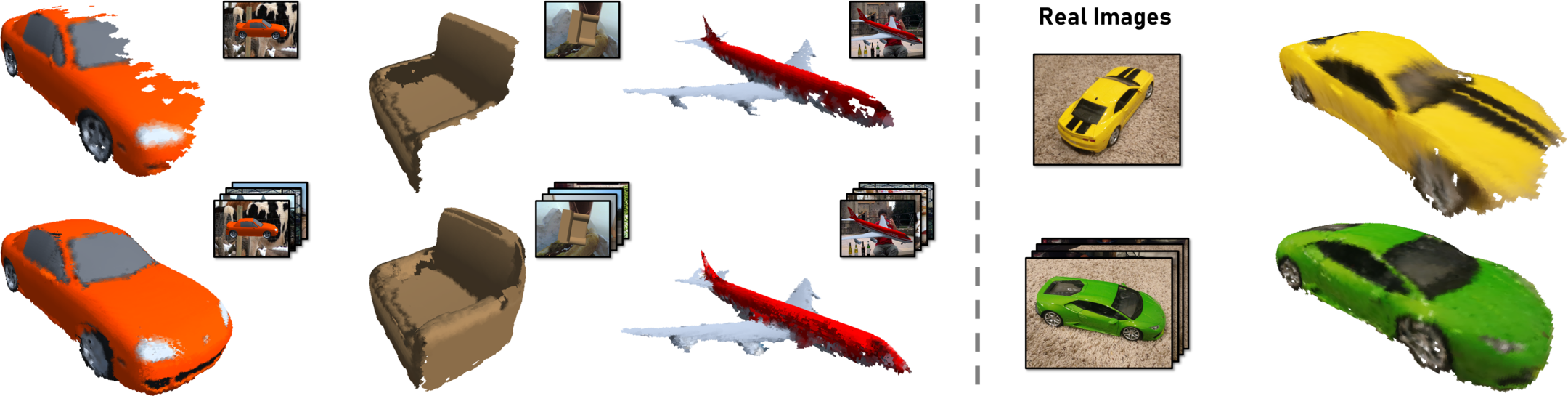}
    \caption{
    Pix2Surf learns to generate a continuous parametric 3D surface of an object seen in one or more views.
    Given a single image, we can reconstruct a continuous partial 3D shape (top row).
    When multiple views are available, we can aggregate the views to form a set of consistent 3D surfaces (bottom row).
    Our reconstructions preserve 2D pixel to 3D shape correspondence that allows the transport of textures, even from real images (last column).
    }
    \label{fig:teaser}
\end{figure}

In this paper, we present \textbf{Pix2Surf}, a method that learns to reconstruct \emph{continuous} and \emph{consistent} 3D surface from single or multiple views of novel object instances, while preserving accurate 2D--3D \emph{correspondences}.
We build upon recent work on category-specific shape reconstruction using \emph{Normalized Object Coordinate Space (NOCS)}~\cite{Wang:2019,Srinath:2019}, which reconstructs the 3D point cloud as a \emph{NOCS map} -- an object-centered depth map -- in a canonical space that is in accurate correspondence with image pixels. Importantly, NOCS maps do not require knowledge of camera geometry.
However, these maps do not directly encode the underlying surface of the object.
In this paper, we present a method that incorporates a representation of the underlying surface by predicting a continuous parametrization that maps a learned \emph{$UV$ parameter space} to 3D NOCS coordinates, similar in spirit to AtlasNet~\cite{Groueix:2018}.
Unlike AtlasNet, however, our approach also provides accurate 2D--3D correspondences and an emergent learned chart that can be used to texture the object directly from the input image.


When multiple views of an object are available, we also present a version of Pix2Surf that is capable of reconstructing an object by predicting an atlas, \ie~view-specific charts assembled to form the final shape.
While in the NOCS approach~\cite{Srinath:2019} individual views can also be directly aggregated since they live in the same canonical space, this na\"ive approach can lead to discontinuities at view boundaries.
Instead, for view-consistent reconstruction, we aggregate multiple views at the feature level and explicitly enforce consistency during training.

Extensive experiments and comparisons with previous work show that Pix2Surf is capable of reconstructing high-quality shapes that are consistent within and across views.
In terms of reconstruction error, we outperform state-of-the-art methods while maintaining the 3C properties.
Furthermore, accurate 2D--3D correspondences allow us to texture the reconstructed shape with rich color information as shown in Fig.~\ref{fig:teaser}.
In summary, the primary contributions of our work are:
%
\begin{packed_itemize}
\item a method to generate a set of \textbf{continuous} parametric 3D surfaces representing the shape of a novel object observed from single or multiple views; 
\item the unsupervised extraction of a learned $UV$ parametrization that retains accurate 2D to 3D surface point \textbf{correspondences}, allowing lifting of texture information from the input image; and 
\item a method to \textbf{consistently} aggregate such parametrizations across different views, using multiple charts.
\end{packed_itemize}



\parahead{Emergent Properties}
A notable emergent property of our network is that the learned $UV$ parametrization domains are consistent across different views of the same object (\ie~corresponding pixels in different views have similar $UV$ coordinates) -- and even across views of related objects in the same class.
This is despite the $UV$ domain maps only being indirectly supervised for consistency, through 3D reconstruction.

\parahead{Scope}
In this work, our focus is on \emph{continuity, consistency, and 2D image--3D surface correspondences}.
We focus on the case when the multi-view images have little overlap, a setting where traditional stereo matching techniques fail.
Our method only requires supervision for the input views and their corresponding NOCS maps but does not require camera poses or ground truth UV parametrization. 
We note that the generated surfaces need not be watertight, and continuity at the seams between views is not guaranteed.


%% file: content/text/02_relwork.tex
\section{Related Work}
%
There is a large body of work on object reconstruction which we categorize broadly based on the underlying shape representation.

\parahead{Voxels}
The earliest deep-learning-based methods predict a voxel representation of an object's shape. Many of these methods are trained as generative models for 3D shapes, with a separate image encoder to obtain the latent code for a given image~\cite{Girdhar16b}. Later methods use more efficient data structures, such as octrees~\cite{Tatarchenko:2017:ogn,Wang-2017-ocnn,Richter:2018} to alleviate the space requirements of explicit voxels. Multiple views can also be aggregated into a voxel grid using a recurrent network~\cite{choy20163d}.
%
%
%
Several methods use supervision in the form of 2D images from different viewpoints, rather than a 3D shape, to perform both single-view and multi-view voxel reconstruction~\cite{lsmKarHM2017,Yan:2016:ptn,drcTulsiani17,henzler2018escaping}. These methods usually use some form of a differentiable voxel renderer to obtain a 2D image that can be compared to the ground truth image. The quality gap of these methods to their counterparts that use 3D supervision is still quite large.
Voxels only allow for a relatively coarse representation of a shape, even with the more efficient data representations.
Additionally, voxels do not explicitly represent an object's surface prompting the study of alternative representations.


\parahead{Point Clouds}
To recover the point cloud of an object instance from a single view, methods with 3D supervision~\cite{Fan:2017,lin2018learning} and without 3D supervision~\cite{insafutdinov18pointclouds} have been proposed. These methods encode the input image into a latent code thus losing correspondences between the image pixels and the output points. Some methods establish a coarse correspondence implicitly by estimating the camera parameters, but this is typically inaccurate.
A recent method reconstructs a point cloud of a shape from multiple views~\cite{Chen:2019:stereo}, but requires ground truth camera parameters.
A large body of monocular or stereo depth estimation methods obtain a point cloud for the visible parts of the scene in an image, but do not attempt to recover the geometry of individual object instances in their local coordinate frames~\cite{bhoi2019monocular}.
%
%
%
NOCS~\cite{Wang:2019,Srinath:2019} obtains exact correspondences between 2D pixels and 3D points by predicting the 3D coordinates of each pixel in a canonical coordinate frame.
NOCS can even be extended to reconstruct unseen parts of an object~\cite{Srinath:2019} (X-NOCS).
All these approaches that output point clouds do not describe the connectivity of a surface, which has to be extracted separately -- a classical and difficult geometry problem.
We extend NOCS to directly recover continuous surfaces and consistently handle multiple views. 
%


\parahead{Implicit Functions}
Poisson Surface Reconstruction~\cite{kazhdan2006poisson,kazhdan2013screened} has long been the gold standard for recovering an implicit surface from a point cloud. More recently, data-driven methods have been proposed that model the implicit function with a small MLP~\cite{chen2018implicit_decoder,Park:2019,Mescheder:2019:Occupancy}, with the implicit function representing the occupancy probability or the distance to the surface. These methods can reconstruct an implicit function directly from a single image, but do not handle multiple views and do not establish a correspondence between pixels and the 3D space.
PiFU~\cite{saito2019pifu} and DISN~\cite{Xu:2019} are more recent methods that establish a correspondence between pixels and 3D space and use per-pixel features to parameterize an implicit function. Both single and multiple views can be handled, but the methods either require ground truth camera poses as input~\cite{saito2019pifu}, or use a network to get a coarse approximation of the camera poses, giving only approximate correspondences~\cite{Xu:2019}.
\jiahui{Some recent works integrate the neural rendering with deep implicit functions~\cite{niemeyer2020differentiable,liu2019learning}, but they depend on the known camera information.}
Furthermore, to obtain an explicit surface from an implicit function, an expensive post-processing step is needed, such as Marching Cubes~\cite{lorensen1987marching} or ray tracing.

\parahead{Parametric Surfaces or Templates}
Several methods attempt to directly reconstruct a parametric representation of a shape's surface. These parametric representations range from class-specific templates~\cite{cmrKanazawa18,kulkarni2019csm}, general structured templates~\cite{Genova:2019:sif}, or more generic surface representations, such as meshes or continuous functions.
%
%
%
%
Pixel2Mesh and its sequel~\cite{Wang:2018:pixel2mesh,Wen:2019:pixel2mesh++} deform a genus-zero mesh based on local image features at each vertex, obtained by projecting the vertices to the image plane(s). Camera parameters are assumed to be known for this projection. 3DN~\cite{wang20193dn} deforms a given source mesh to approximate a single target image, using global features for both the source and the target, without establishing correspondences to the target pixels.
Several methods use 2D images instead of 3D meshes as supervisory signal~\cite{kato2018renderer,liu2019softras,petersen2019pix2vex,kato2019vpl} using differentiable mesh renderers. This makes it easier to collect training data, but the accuracy of these methods still lags behind methods with 3D supervision.
%
AtlasNet~\cite{Groueix:2018} represents shapes with continuous 2D patches that can be inferred from a single input image, or from a video clip~\cite{lin2019photometric}. \jiahui{Mesh DeformNet~\cite{pan2019deep} introduces topology modification to AtlasNet.}
Similar to AtlasNet, we use a 2D patch as a $UV$ parametrization, but we handle multiple non-adjacent views and establish correspondences between 2D pixels and 3D surface points.

%% file: content/text/03_overview.tex
\section{Preliminaries}
%
We build our approach upon two previous ideas that we describe below.
%
%

\begin{wrapfigure}{r}{0.4\textwidth}
  \centering
    \includegraphics[width=0.37\textwidth]{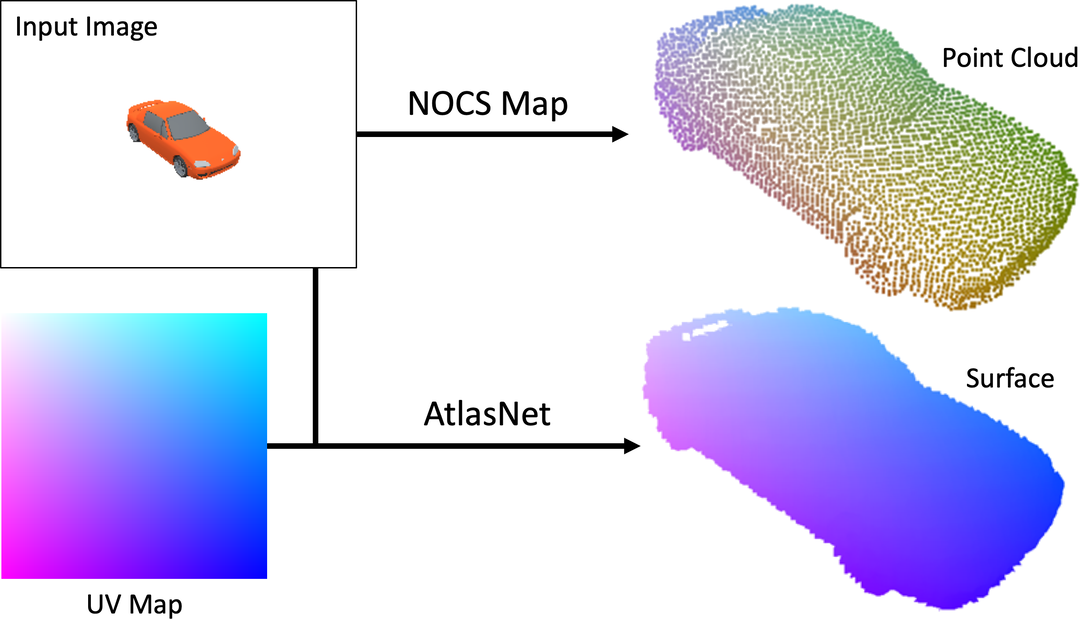}
    \caption{Given a single image, \mbox{X-NOCS}~\cite{Srinath:2019} reconstructs a point cloud preserving pixel--3D correspondece. AtlasNet~\cite{Groueix:2018} learns shape as a continuous surface.}
    \label{fig:prelim}
\end{wrapfigure}

\parahead{(X-)NOCS}
Normalized object coordinate space (NOCS) is a canonicalized unit
container space used for category-level reasoning of object pose, size, and shape~\cite{Srinath:2019,Wang:2019}.
Instances from a given object category are \emph{normalized} for their position, orientation, and size, thus disentangling intra-category shape variation from the exact pose and size of instances.
NOCS maps (see Fig.~\ref{fig:prelim}) are perspective projections of the 3D NOCS shape onto a specific camera and can be interpreted as \textbf{object-centered depth maps} that simultaneously encode mask and partial shape of the object.
When used to predict 3D point cloud from images, NOCS maps retain correspondences from 2D pixels to 3D points, and can be used to transport image texture directly to 3D.
X-NOCS is an extension of NOCS maps to also encode the occluded parts of a shape~\cite{Srinath:2019}.
However, using NOCS maps for reconstruction results in a discontinuous point cloud.

\parahead{Surface Parametrization}
%
A two-manifold surface in 3D can be mapped to a 2D plane (\emph{chart}) parametrized by two coordinates $(u, v)$.
This $UV$ parametrization of a 3D surface is widely used in computer graphics and, more recently, in 3D shape reconstruction~\cite{Groueix:2018,kulkarni2019csm}.
The parameterization can be limited in expressing complex shapes, depending on the functional formulation used.\setlength{\intextsep}{1pt}
For example, in typical CAD settings, low-degree polynomial or rational functions are used to represent the mappings.
In our case, instead, we use a fully connected network to overcome the limitation of expressibility.
A single map, however, still lacks the ability to describe complicated shapes with high-genus topology. 
Thus, multiple charts are often used, where multiple 2D planar patches are mapped by separate maps to a 3D surface -- effectively partitioning the surfaces into parts, each of which is the image of a different map in the chart.
We show how a single chart can be used for 3D shape reconstruction while multiple charts allow consistent reconstruction over multiple views while still preserving pixel to 3D correspondence.

%% file: content/text/04_method.tex
\begin{figure*}[t!]
    \centering
   \includegraphics[width=\textwidth]{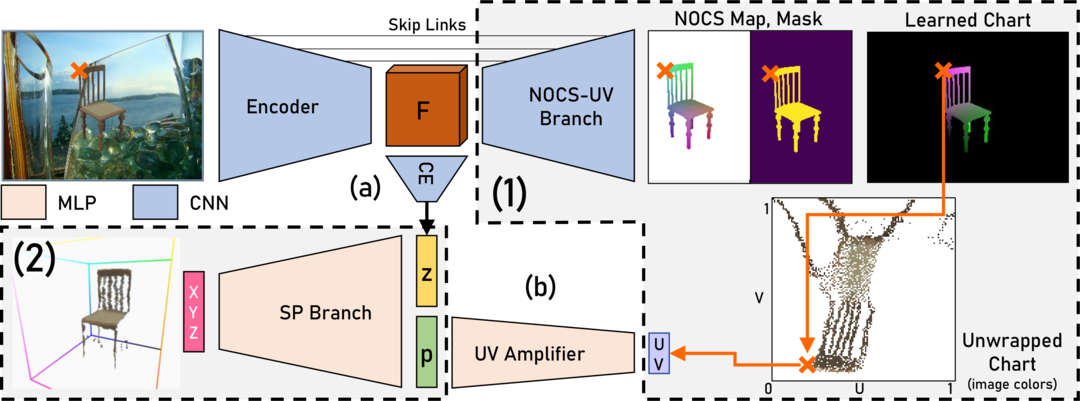}
    \caption{Single-View Single-chart Pix2Surf network architecture.
    The input image is processed using an encoder-decoder architecture to predict a NOCS map, object mask, and a learned chart (top right \jiahui{two channels, color coded}).
    The Surface Parameterization branch takes sampled and amplified chart coordinates $\mathbf{p}$ and a latent image code $\mathbf{z}$ to predict the final continuous surface.
    Unwrapped chart \jiahui{(bottom right)} refers to a visualization \jiahui{of foreground colors using the predicted two-channel learned chart (top right) as coordinate}. 
    The colors of the input image can be transported to all intermediate steps ($\boldsymbol{\textcolor{orange}{\times}}$ and arrows).
    }
    \label{fig:single}
\end{figure*}
\section{Pix2Surf: Pixels to Surface}
%
Our goal is to predict a continuous and consistent parametric 3D surface for a novel object instance observed from one or more views.
\jiahui{The word ``continuous'' parametric surfaces in our method refers to parametric $C^0$ continuity (similar to AtlasNet~\cite{Groueix:2018}), \ie~any continuous trajectory over the chart space maps to a continuous curve in 3D space.}
Additionally, we would like to preserve correspondences between 2D pixels and 3D surface points.
We first describe our approach for reconstructing a 3D surface from a single image using a single chart, and then generalize it to multiple views using an atlas.

\subsection{Single-View Single-Chart Pix2Surf}
\label{sec:single_view}
%
At inference time, the single-view version of Pix2Surf takes an RGB image of an object observed from an arbitrary camera as input.
We use a CNN to extract image features that compactly encode object shape.
The features are then processed by two branches: (1)~the \textbf{NOCS-UV branch} is a CNN that \jiahui{estimates a mask, a learned UV map, and a NOCS map} and (2)~the \textbf{Surface Parametrization (SP) branch} is an MLP that generates a continuous 3D surface.
This single-view, single-chart architecture is shown in Fig.~\ref{fig:single}.

\parahead{(1) NOCS-UV Branch}
Similar to X-NOCS~\cite{Srinath:2019}, we predict the NOCS map and mask that encode the partial shape of the object observed in the image.
We use an encoder-decoder architecture building on top of SegNet~\cite{badrinarayanan2017segnet} and VGG~\cite{simonyan2014very}.
Our network uses skip connections and shares pool indices between the encoder and the decoder.
The predicted NOCS maps and masks are the same size as the input image.
During training, the object mask is supervised with a binary cross entropy loss and the NOCS map is supervised with an $L^2$ loss.
Note that the NOCS map here is \textbf{\emph{not}} our final 3D output, but acts as an intermediate supervision signal for the network.

\begin{wrapfigure}{r}{0.6\textwidth}
  \centering
    \includegraphics[width=0.6\textwidth]{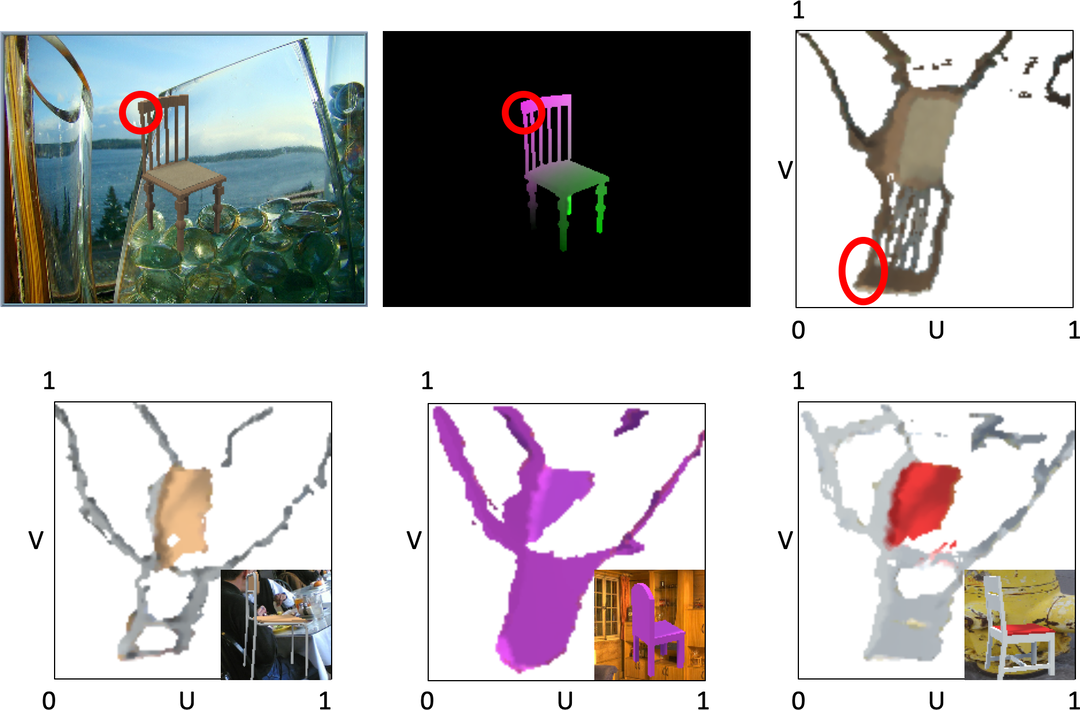}
    \caption{Given an object image (row 1, col 1), our network predicts a 2-channel image without explicit supervision (row 1, col 2, color coded).
    Remarkably, the output of these two channels visualized in a $UV$ space (row 1, col 3) show that the network has learned to unwrap the 3D shape onto a plane (corresponding patches shown in red). This unwrapping is consistent over multiple views, and even across multiple object instances (last row). \jiahui{For more unwrapped charts of cars and airplanes please see supplementary Fig.~S.3.}}
    \label{fig:learnedUV}
\end{wrapfigure}

%
\parahead{Emergence of a Chart}
Different from previous work, we predict a 2-channel output in additional to the NOCS map and mask.
These 2 channels are not explicitly supervised during training, so the network can predict any value between 0 and 1.
However, when jointly trained with the other branches, we observe the \textbf{emergence} of a \emph{learned chart} in these 2 channels (see Fig.~\ref{fig:learnedUV}).
The network discovers how to unwrap an object shape onto a flat surface.
\textbf{Remarkably, this learned chart is (almost) consistent across multiple views and even across instances}.
During reconstruction, each image pixel's learned chart coordinates are passed on to the SP branch.
We show that using the learned chart coordinates is superior to using arbitrary UV coordinates like AtlasNet~\cite{Groueix:2018}, or alternatively using the original image coordinates (Image2Surf, Sec.~\ref{sec:visible_reconstruction}).
Additionally, we preserve exact correspondences between input image pixels and the learned chart.

\setlength{\intextsep}{1pt}
\parahead{(a) Code Extractor (CE)}
We use a small CNN to reduce the high dimensional feature map extracted by the encoder to make a more compact global code for the SP branch.
This CNN contains two convolutional layers (512 and 1024 output channels), batch normalization, and ELU activation.
The output is a latent code $\mathbf{z}$ of size 1024 and is passed to the SP branch.

\parahead{(b) UV Amplifier}
Before we use the learned chart coordinates as an input to the SP branch, we process each UV coordinate with a \emph{UV amplifier} MLP.
The motivation for this comes from the information imbalance the two inputs to the SP branch -- one input is the global latent code $\mathbf{z}$  which has 1024 dimensions, while the UV coordinates would have only 2 dimensions.
To overcome this, we \emph{amplify} the UV coordinates to $\mathbf{p}$ (256 dimensions) using a 3-layer MLP that progressively amplifies the 2 coordinates (2, 64, 128, 256).
This allows the SP branch to make use of the image and UV information in a more balanced manner.

\parahead{(2)~SP Branch}
Similar to AtlasNet~\cite{Groueix:2018}, our surface parametrization (SP) branch takes the global latent code $\mathbf{z}$ from the code extractor (CE) and the amplified coordinates $\mathbf{p}$ as input and produces a continuous 3D position as the output.
Note that the learned chart coordinates can be continuously sampled at inference time.
\jiahui{The continuity of the output 3D surface emerges from our use of a continuous MLP mapping function between the uv coordinates and the output 3D positions~\cite{Groueix:2018}.}
Our SP branch is a MLP with 9 layers and skip connection every 2 layers (input: 1024+256, intermediate: 512, last: 3).
Since we train on canonically oriented ShapeNet models, the predicted 3D surface also lies within the canonical NOCS container~\cite{Wang:2019}.

Our approach has three key differences to AtlasNet.
First, we use a UV amplifier to transform the 2D $UV$ coordinates to higher dimensions allowing better information balancing.
Second, the learned chart is in direct correspondence with the pixels of the input image (see Fig.~\ref{fig:learnedUV}).
This allows us to transport appearance information directly from the image to the 3D surface.
Third, our sampling of the $UV$ chart is learned by a network (NOCS-UV branch) instead of uniform sampling, which enables us to reconstruct complex topologies.
Our inference processing allows us to sample any continuous point in the learned chart space within the predicted object mask allowing the generation of \textbf{continuous textured 3D surface}.

\parahead{Training}
The encoder and decoder CNNs are first initialized by training them on the NOCS map and mask prediction tasks using only the $L^2$ and BCE losses.
Subsequently, we jointly train the NOCS-UV and SP branches, code extractor, and UV amplifier end-to-end.
The joint loss is given as,
%
\begin{equation}
\mathcal{L}_I = w_1 \, (w_n \, \mathcal{L}_n + w_m \, \mathcal{L}_m) + w_2 \, \mathcal{L}_s,
\end{equation}
where $\mathcal{L}_n$ and $\mathcal{L}_m$ are the $L^2$ NOCS map and BCE losses respectively, $w_n, w_m$ are the weights for the NOCS map and mask prediction respectively, and $w_1, w_2$ are the weights for the NOCS-UV and SP branches respectively.
For the SP branch we supervise on K points sampled randomly from within the foreground mask.
For each sampled point, a corresponding amplified chart coordinate $\mathbf{p}$ is predicted without any supervision.
This is concatenated with the global latent code $\mathbf{z}$ to predict the final 3D surface position.
Empirically, we found the best hyperparameters to be: $K = 4096, w_1 = 0.1, w_2 = 0.9, w_n = 0.7, w_m = 0.3$.
The loss for the SP branch is given as, $L_{s} = \frac{1}{K} \, \sum_{i=1}^{K}{\|\mathbf{x}_i - \hat{\mathbf{x}}_i\|_2}$, 
%
where $\mathbf{x}$ and $\hat{\mathbf{x}}$ are the ground truth and predicted 3D surface position obtained from the 3D ShapeNet models (same as ground truth NOCS map values).
During inference, we can predict a continuous 3D surface for any given image and its learned chart coordinate.
Please see the supplementary document for more details on inference and final 3D model generation.
\newpage

%% file: content/text/04_1_multi.tex
\subsection{Multi-View Atlas Pix2Surf}
\label{sec:multi_view}
\begin{figure*}[!b]
    \centering
    \includegraphics[width=0.99\textwidth]{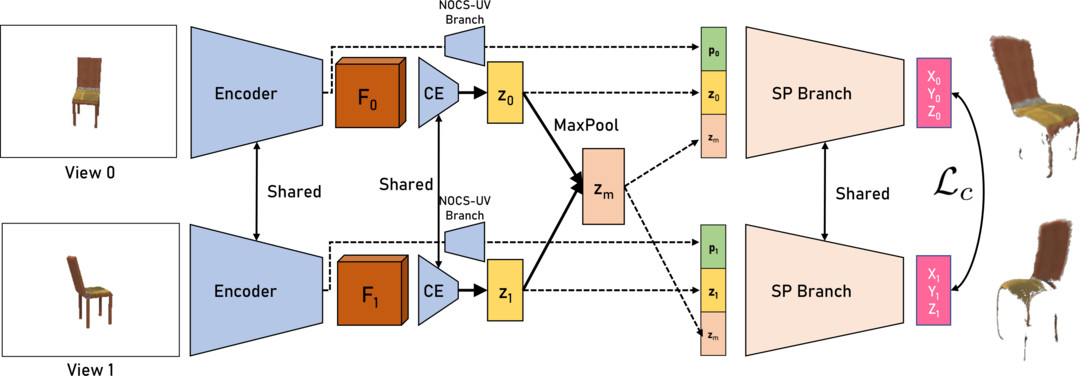}
    \caption{Multi-view atlas network architecture.
    The multi-view network allows multiple charts to be consistently aggregated.
    This network has two main features: (1) the MaxPool operation to pool features between views, and (2) a multi-view consistency loss $\mathcal{L}_C$ that ensures corresponding points produce 3D surface points that are nearby.
    Only two views are shown in this figure, but we use multiple views during training and inference.
    The encoder, NOCS-UV branch, CE branch, and SP branches share weights.}
    \label{fig:multi}
\end{figure*}

\setlength{\intextsep}{1pt}
\begin{wrapfigure}{r}{0.5\textwidth}
  \centering
    \includegraphics[width=0.5\textwidth]{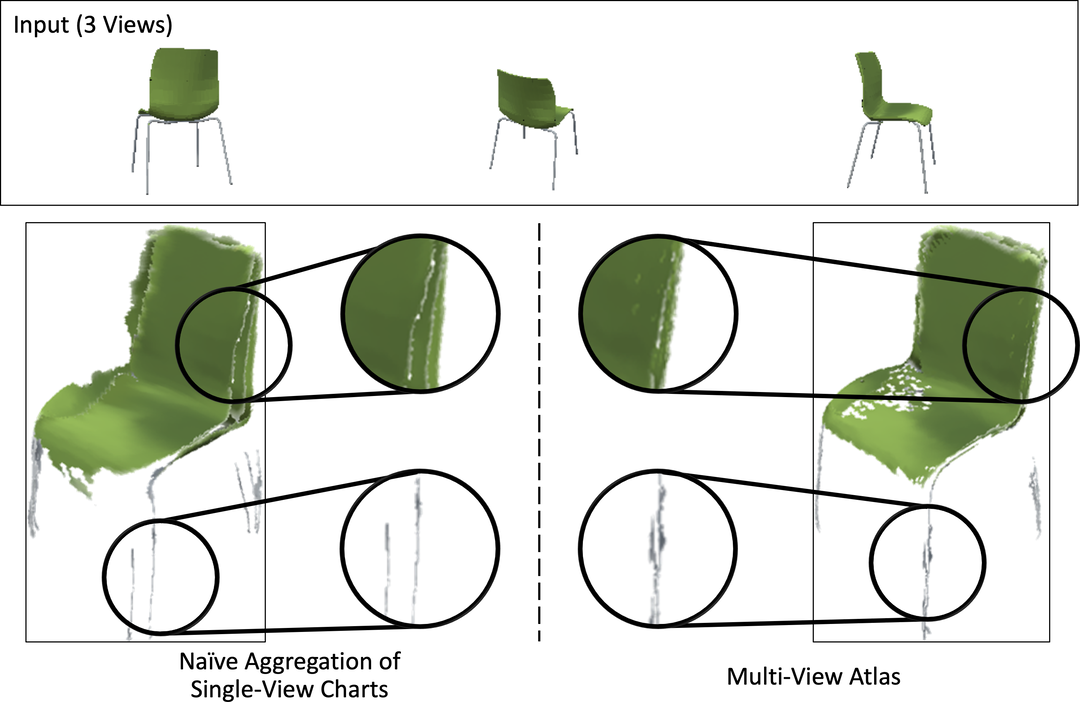}
    \caption{
    Given 3 views, na\"ive aggregation of individual charts leads to discontinuities or double surfaces (left).
    Our multi-view atlas method produces more consistent surfaces (right), for instance, at the legs and backrest.
    }
    \label{fig:naive}
\end{wrapfigure}
The method described above is suitable when we have a single view of the object.
When multiple views are available, we could naively extend the single view network and combine the generated surfaces using a union operation.
However, this leads to sharp discontinuities (Fig.~\ref{fig:naive}).
To overcome this issue, we propose a generalization of our single-view single-chart method to consistently aggregate 2D surface information from multiple views, using an \emph{atlas} \ie~a separate learned chart (UV map) for each view.
Fig.~\ref{fig:multi} shows an overview of our multi-view network.
This design shares similarities with the single view network but has additional multi-view consistency which is enforced both at the feature level through a feature pooling step, and using a consistency loss for better 3D surface generation.
\parahead{Multi-View Feature Pooling}
The goal of this step is to promote multi-view information sharing at the \emph{feature level} (see Fig.~\ref{fig:multi}).
Different from the single view network, the 
latent codes $\mathbf{z}_i$ extracted for each view $i$ (using a shared encoder and code extractor) are maxpooled into a common shared multi-view latent code $\mathbf{z}_m$.
Intuitively, this shared latent code captures the most salient information from each view.

\parahead{Atlas}
Similar to the single view network, we learn a chart for each view.
The chart coordinates for each view $\mathbf{p}_i$ are extracted using the NOCS-UV branch with weights shared between the views.
Although the NOCS-UV branch weights are shared, one chart is predicted for each view  -- thus, we have an atlas.
Note that the network is free to predict different chart coordinates for each view.
However, we observe that similar parts of objects in different images map to similar locations on their respective charts (see Fig.S3 in supplementary document).
This indicates that our network is \textbf{discovering the notion of image \jiahui{cross-view} correspondence \jiahui{(note that this is different from 2D-3D correspondence)}}. 
As in the single-view network, chart coordinates are passed through a shared UV amplifier.

We concatenate the shared latent code $\mathbf{z}_m$ to each of the per-view latent codes $\mathbf{z}_i$.
This concatenated multi-view code and the learned per-view chart coordinates $\mathbf{p}_i$ are passed to the SP branch.
The UV amplifier, code extractor and structure of the learned UV map are similar to the single view network.

\parahead{Multi-View Loss}
In addition to the $L^2$ loss function on the 3D surface generated by the SP branch, we also have a multi-view consistency loss.
This loss enforces corresponding points on multiple views to predict similar 3D surface positions.
To obtain correspondence information at training time, we sample a random set of foreground points within the mask and find the exact match of the ground truth NOCS values of that pixel in the other input views.
Note that this correspondence information is \textit{not} provided as additional supervision -- the ground truth NOCS maps already contain this information since corresponding points multiple views have the same NOCS position.
Given these correspondences, the multi-view consistency loss for a pair of views is given as, \mbox{$\mathcal{L}_C = \frac{1}{|\mathcal{P}|}\sum_{(i,j)\in\mathcal{P}}\|\mathbf{x}_i-\mathbf{x}_j\|_2$}, 
where $\mathbf{x}_{i,j}$ are the paired predicted xyz from two different views and the set $\mathcal{P}$ contains all matched correspondence pair from \jiahui{these} two views.
During training, within each mini-batch, we sample multiple views per object and compute the loss for all possible pairs.

\parahead{Training}
The multi-view network is trained similar to the single view model.
The NOCS-UV branch is first trained and subsequently the whole network is trained end-to-end.
The loss function we use is $\mathcal{L}_M = \mathcal{L}_I + \frac{w_3}{a} \, \sum_{j = 0}^{a} \, \mathcal{L}_C$, where $a$ denotes the number of pairs of views within that batch, and $w_3$ is the correspondence loss weight empirically set to 0.9.
\jiahui{We set $w_n,w_m$ to 0.1 inside $\mathcal{L}_I$.}
Please see the supplementary document for more details on inference and final 3D model generation.

%% file: content/text/05_experiments_new.tex
\section{Experiments}
%
We present extensive experimental comparison of Pix2Surf with several recent single- and multi-view reconstruction methods, and validate our design choices.
We do so by focusing on the 3C properties (consistency, correspondence and continuity) for visible surface reconstruction (Sec.~\ref{sec:visible_reconstruction}). 
Since we learn a strong prior over shapes, we can also estimate surfaces that are hidden in the input image (Sec.~\ref{sec:hidden_generation}).
\minhyuk{For the details of training, inference, and evaluation metrics, and also ablations, more comparisons, and results with real images, refer to the supplementary document.}

\parahead{Dataset}
For quantitative comparisons, we use ShapeNetPlain~\cite{Srinath:2019} dataset which consists of 5 random views for each shape in ShapeNet~\cite{chang2015shapenet} with a white background.
For additional robustness to the background found in real-world images, we train Pix2Surf on ShapeNetCOCO~\cite{Srinath:2019} which consists of $20$ random views of each ShapeNet object with a random background from MS COCO~\cite{lin2014microsoft}.
We use this dataset for all qualitative results and for real-world results. \jiahui{Each shape category is trained separately in all experiments.}


\parahead{Experimental Setting}
We follow the experimental setup of X-NOCS~\cite{Srinath:2019}.
Our ground truth for each input image is the point cloud represented by the NOCS map (or X-NOCS map for hidden surface) provided in the dataset~\cite{Srinath:2019}.
The outputs  of all methods are converted to a NOCS map (using the ground truth camera pose) allowing us to compute metrics even for partial shapes.
Multi-view experiments use all 5 views in the dataset to reconstruct a surface, using the same dataset as the single-view experiments.
All metrics are computed per-view and then averaged up, making the single- and multi-view values comparable in our quantitative experiments. 

\newcommand{\rmetric}[1]{{\textcolor{red}{\textbf{#1}}}}
\newcommand{\corr}[1]{{\textcolor{OliveGreen}{\textbf{#1}}}}
\newcommand{\cons}[1]{{\textcolor{blue}{\textbf{#1}}}}
\newcommand{\cont}[1]{{\textcolor{cyan}{\textbf{#1}}}}
\parahead{Metrics}
We quantify the quality of reconstructed surfaces with several metrics.
The \rmetric{reconstruction error} of predictions is computed as the Chamfer distance~\cite{Barrow:1977:Chamfer,fan2017point} between the estimated surface and the ground truth NOCS map (interpreted as a point cloud).
To obtain points on a reconstructed surface, we convert it into a NOCS map using the ground truth camera pose.
%

In addition to the accuracy of reconstructed surfaces, we quantify the 3C properties of a surface with the following metrics.
The 2D--3D \corr{correspondence error} measures the accuracy of the estimated correspondence between input pixels and 3D points on the reconstructed surface.
The error for each foreground pixel is the distance between the estimated 3D location of the pixel and the ground truth location.
Unlike the Chamfer distance, this uses the 2D--3D correspondence to compare points.
We average over all foreground pixels to obtain the correspondence error of a surface.
The \cons{multi-view consistency error} was defined in Sec.~\ref{sec:multi_view} as the 3D distance between corresponding points in different views.
We average the distance for a given point over all pairs of views that contain the point.
Corresponding points are found based on the ground truth NOCS map.
%
%
We measure \cont{discontinuity} based on the surface connectivity.
\jiahui{While the continuity of Pix2Surf is induced by our use of a continous MLP as mapping from uv space to 3D space~\cite{Groueix:2018}, the mapping from the \emph{input image} to the 3D space should \emph{not} be $C^0$-continuous everywhere, due to self occlusions and boundaries of the 3D shape.} The reconstructed surface should have the same $C^0$ discontinuities as the ground truth surface. We define a $C^0$ discontinuity as large difference in the 3D locations of the neighboring pixels in a NOCS map (above a threshold of 0.05).
We take a statistical approach to measure the surface connectivity, by computing a histogram over the 3D distances between neighboring pixels that are discontinuous.
The \jiahui{dis}continuity score is the correlation of this histogram to a histogram of the ground truth surface. A higher score indicates a distribution of discontinuities that is more similar to the ground truth surface. \jiahui{Note that continuity is a property induced from method design itself, and the score can penalize the over-smooth case from methods that produces continuous prediction.} 


\begin{table}[t]
\setlength{\tabcolsep}{0.5pt}
\centering
\relsize{-1}
\caption{Visible surface reconstruction. We compare our method to a baseline and three state-of-the-art methods evaluating reconstruction accuracy and the 3C properties.
The top half of the table shows single-view reconstruction, the bottom half is multi-view reconstruction.
Note how Pix2Surf is close to the top performance in each of the metrics, while all other methods have significant shortcomings. \jiahui{The \rmetric{Recons.~Error} and \corr{Correspond.~Error}, \cons{Consistency Error} are all multiplied by $10^3$.}
}
\label{tbl:visible_recon}
{
\begin{tabular}{@{}lccccccccccccccccccc@{}} 
\toprule
& \multicolumn{4}{c}{\rmetric{Recons.~Error} $\downarrow$} & \phantom{.} & \multicolumn{4}{c}{\corr{Correspond.~Error} $\downarrow$} & \phantom{.} & \multicolumn{4}{c}{\cons{Consistency Error} $\downarrow$} & \phantom{.} & \multicolumn{4}{c}{\cont{Disconti. Score} $\uparrow$} \\
\cmidrule{2-5} \cmidrule{7-10} \cmidrule{12-15} \cmidrule{17-20}
& car & chair & plane & \textbf{avg.} &
& car & chair & plane & \textbf{avg.} &
& car & chair & plane & \textbf{avg.} &
& car & chair & plane & \textbf{avg.} \\
\midrule
Im.2Surf & 2.23 & 3.81 & 2.66 & 2.90 && \textbf{8.49} & 9.54 & 8.76 & 8.93 && 13.08 & 12.55 & 10.75 & 12.13 && 0.46 & 0.39 & 0.35 & 0.40 \\
X-NOCS & 2.25 & 2.95 & 2.08 & 2.43 && 12.82 & 8.63 & 8.93 & 10.13 && 18.93 & 12.00 & 10.59 & 13.84 && 0.59 & \textbf{0.47} & 0.59 & 0.55 \\
AtlasNet & \textbf{1.54} & 3.36 & 3.15 & 2.68 && -- & -- & -- & -- && -- & -- & -- & -- && 0.68 & 0.39 & 0.64 & 0.57 \\
\textbf{Pix2Surf} & 1.67 & \textbf{1.91} & \textbf{1.61} & \textbf{1.73} && 9.52 & \textbf{5.79} & \textbf{7.19} & \textbf{7.50} && \textbf{12.72} & \textbf{7.75} & \textbf{8.48} & \textbf{9.65} && \textbf{0.69} & 0.43 & \textbf{0.65} & \textbf{0.59} \\
\midrule
X-NOCS & 2.89 & 2.80 & 2.19 & 2.63 && 14.30 & 9.48 & 8.95 & 10.91 && 22.18 & 14.26 & 11.65 & 16.03 && \textbf{0.67} & \textbf{0.48} & 0.54 & 0.56 \\
P2M++ & 2.88 & 5.59 & 3.24 & 3.90 && -- & -- & -- & -- && -- & -- & -- & -- && 0.67 & 0.36 & 0.63 & 0.55 \\
\textbf{Pix2Surf} & \textbf{1.41} & \textbf{1.78} & \textbf{1.38} & \textbf{1.52} && \textbf{8.49} & \textbf{5.84} & \textbf{7.06} & \textbf{7.13} && \textbf{10.98} & \textbf{6.65} & \textbf{7.50} & \textbf{8.38} && 0.66 & 0.43 & \textbf{0.66} & \textbf{0.58} \\
\bottomrule
\end{tabular}
}
\end{table}

\subsection{Visible Surface Reconstruction}
\label{sec:visible_reconstruction}
We compare the quality of single- and multi-view reconstructions to one baseline [Image2Surf (single-view)], and three state-of-the-art methods [AtlasNet~\cite{Groueix:2018} (single-view), X-NOCS~\cite{Srinath:2019} (single- and multi-view), Pixel2Mesh++~\cite{Wen:2019:pixel2mesh++} (multi-view)].
%
Note that Pix2Surf deals with a more challenging problem compared to AtlasNet and Pixel2Mesh++: (1)~we predict 2D--3D correspondences (AtlasNet does not), and (2)~we do not require camera geometry information as input (Pixel2Mesh++ does).
In this section, we only focus on reconstructing \emph{visible} surfaces, but we also report hidden surface generation in the next section.

The single-view performance of each method in all of our metrics is shown in the first four rows of Table~\ref{tbl:visible_recon}, and the multi-view performance in the last three rows. Metrics are comparable across single- and multi-view methods. For each of the four metrics, we show the performance on each dataset category, and an average over all categories.

\parahead{Image2Surf}
This baseline is similar to Pix2Surf, but takes image $UV$ coordinates (normalized by predicted mask) as input to the UV amplifier instead of the learned $UV$ chart, \ie~the input image is the chart.
We observe that it is hard for the network to learn depth discontinuities, resulting in over-smoothed occlusion boundaries (see supplementary document).
The over-smoothing is reflected in a high reconstruction error, and particularly low \jiahui{discontinuity correlation score}.
This comparison justifies our design to include a learned $UV$ chart.

\begin{wrapfigure}{r}{0.5\textwidth}
  \centering
    \includegraphics[width=0.5\textwidth]{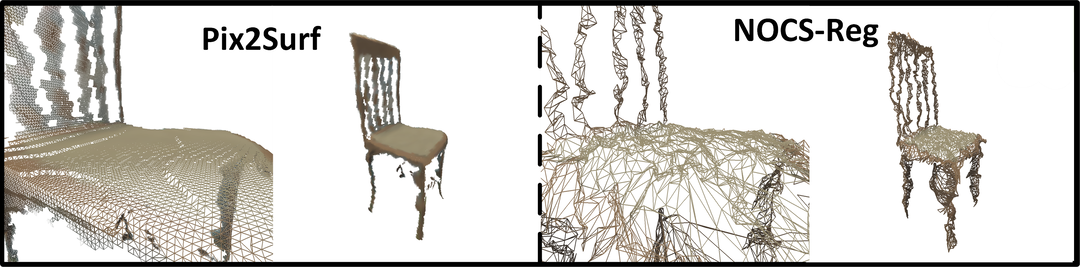}
\caption{Our results (left) compared with surface-agnostic X-NOCS (right), visualized with image connectivity. Pix2Surf produces significantly smoother results.}
\label{fig:baseline}
\end{wrapfigure}

\parahead{X-NOCS}
This is a state-of-the-art reconstruction method that predicts a 3D point cloud, \ie~a 3D point for each foreground pixel.
Since X-NOCS has no notion of surface connectivity, there is no coordination between neighboring points, resulting in poor reconstruction accuracy and noisy output point clouds (see Fig.~\ref{fig:baseline}).
%
\setlength{\intextsep}{1pt}
%
Note that the output point cloud from X-NOCS can capture the right discontinuity. However, it can only produce discrete noisy point cloud instead of continuous surfaces.



\parahead{AtlasNet}
This method also uses an explicit surface parametrization, giving it
a low reconstruction error on the Car category.
However, since the parametrization is not learned and has a fixed layout and connectivity, the reconstruction error increases significantly for categories with more complex shapes and topologies, such as Chair and Airplane.
Correspondence and \jiahui{multi-view consistency}  
are not evaluated, since AtlasNet lacks pixel-to-point correspondences \jiahui{and works only for a single view}.

\parahead{Pixel2Mesh++}
This method deforms a given starting mesh in a coarse-to-fine approach to approximate an object shown from multiple views. In each refinement step, a mesh vertex is deformed based on a small image neighborhood around the projection of the vertex in each view. Unlike in our method, ground truth camera positions need to be known for this projection. The fixed connectivity and topology of the starting mesh results in 
a higher reconstruction error. Since correspondence and multi-view consistency are trivial given a ground truth camera model, we do not evaluate these properties.

Unlike the previous methods, \textbf{Pix2Surf} learns a \jiahui{continuous} parametrization of the surface that does not have a fixed topology or connectivity.
This gives us more flexibility to approximate complex surfaces, for instance, to correctly place holes that can model $C^0$ discontinuities.
This explains our high \jiahui{dis}continuity correlation scores which also benefits the accuracy of reconstruction and 2D-3D correspondence.
%
In the multi-view setting, Pix2Surf shares information across the views, improving the overall reconstruction accuracy.
For example, surfaces that are only visible at a very oblique angle in one view can benefit from additional views.
Our use of a consistency loss additionally ensures an improvement of the multi-view consistency over the baselines, and a lower consistency error compared to single view Pix2Surf (Fig.~\ref{fig:naive}). 
%
We observe that Pix2Surf is the only method that has top performance on all quality metrics (reconstruction and 3C properties), all other methods reconstruct surfaces that fall short in at least some of the metrics.

\begin{table}[t]
\begin{minipage}{.66\linewidth}
\setlength{\tabcolsep}{0.5pt}
\centering
\relsize{-1}
\caption{We compare the reconstruction error of visible and hidden surfaces (trained jointly) for Pix2Surf and X-NOCS [single view (sv.) and multi-view (mv.)].
The learned parametrization of Pix2Surf also benefits from hidden surface generation, and the additional reconstruction of the hidden surface does not adversely affect the accuracy of the visible surfaces.}
\label{tab:hidden_recon}
\newcolumntype{Y}{>{\centering\arraybackslash}X}
\begin{tabularx}{\textwidth}{lYYYYYYYYY}
\toprule
& \multicolumn{4}{c}{\rmetric{Visible Error} $\downarrow$} & \phantom{.} & \multicolumn{4}{c}{\rmetric{Hidden Error} $\downarrow$} \\
\cmidrule{2-5} \cmidrule{7-10}
& car & chair & plane & \textbf{avg}. && car & chair & plane & \textbf{avg}. \\
\midrule
X-NOCS (sv.) & 2.25 & 2.95 & 2.08 & 2.43 && 1.86 & 3.34 & 2.25 & 2.48\\
X-NOCS (mv.) & 2.89 & 2.80 & 2.19 & 2.63 && 3.11 & 3.32 & 2.03 & 2.82\\
\textbf{Pix2Surf} & \textbf{1.66} & \textbf{2.01} & \textbf{1.66} & \textbf{1.78} && \textbf{1.52} & \textbf{2.47} & \textbf{1.77} & \textbf{1.92}\\
\bottomrule
\end{tabularx}
\end{minipage}\hfill%
\begin{minipage}{.32\linewidth}
\centering
\includegraphics[width=\columnwidth]{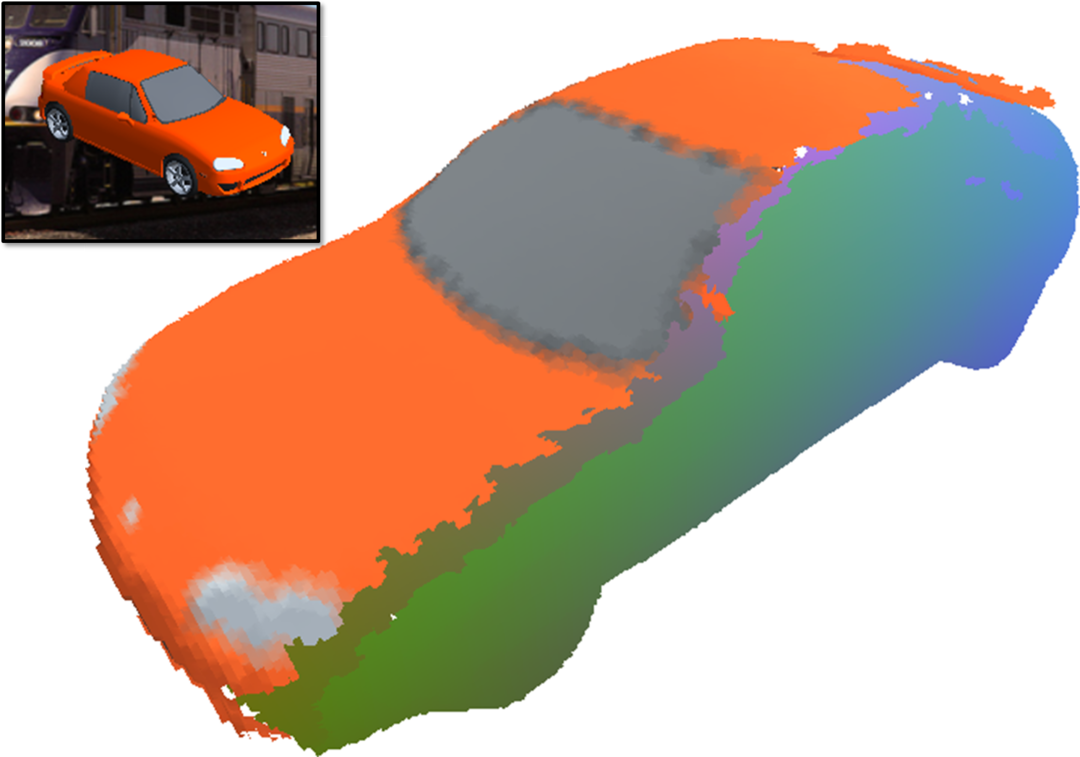}
\captionof{figure}{Pix2Surf can reconstruct both visible (textured) and hidden parts (color coded).}
\label{fig:hidden_surf}
\end{minipage}
\end{table}

\subsection{Hidden Surface Generation}
\label{sec:hidden_generation}
Since Pix2Surf learns a strong prior of the shapes it was trained on, we can generate plausible estimates for surfaces in parts of the object that are not directly visible in the image~\jiahui{(see Fig.~\ref{fig:hidden_surf})}.
Similar to X-NOCS, we represent a 3D object with two layers: a visible layer that we reconstruct in the experiments described previously, and a hidden layer denoting the last intersection of camera rays~\cite{Srinath:2019}. 
Pix2Surf can be easily extended to reconstruct hidden surface farthest from the camera by adding additional output channels to the NOCS-UV branch.
The rest of the architecture remains the same with the learned $UV$ parametrization additionally also learning about the hidden surface.
%
In Table~\ref{tab:hidden_recon}, we show our performance when jointly reconstructing the visible and hidden surfaces from an image. 
We compare to both the single- and multi-view version of X-NOCS on all categories.
The improvement in accuracy for our method shows that hidden surfaces benefits from our learned parametrization as well.
Comparing the performance of the visible surface reconstruction to Table~\ref{tbl:visible_recon}, we see that the joint reconstruction of visible and hidden surfaces does not \jiahui{significantly} decrease the reconstruction accuracy of the visible surfaces.

%% file: content/text/06_conclusion.tex
\section{Conclusion}
We have presented Pix2Surf, a method for predicting 3D surface from a single- or multi-view images. Compared with the previous work, Pix2Surf simultaneously achieves three properties in the prediction: \textbf{continuity} of the surface, \textbf{consistency} across views, and pixel-level \textbf{correspondences} from the images to the 3D shape.
By attaining these properties, our method enables the generation of high-quality parametric surfaces, readily integrating the output surfaces from multi-views, and lifting texture information from images to the 3D shape.
%
In future work, we will explore ways of guaranteeing continuity even across different views and improving the quality of mapped textures. \jiahui{Another interesting direction is to exploit the intermediate learned chart as a container for material properties.}
A longer-term goal would be to investigate how the network can generalize across multiple categories. 

\bigbreak
\parahead{Acknowledgements}
We thank the anonymous reviewers for their comments and suggestions. This work was supported by a Vannevar Bush Faculty Fellowship, NSF grant IIS-1763268, grants from the Stanford GRO Program, the SAIL-Toyota Center for AI Research, AWS Machine Learning Awards Program, UCL AI Center, and a gift from the Adobe.

%% file: content/supp/text/10_supplementary.tex

\ifpaper
  \newcommand\refpaper[1]{\unskip}
\else
  \makeatletter
  \newcommand{\manuallabel}[2]{\def\@currentlabel{#2}\label{#1}}
  \makeatother
  \manuallabel{sec:single_view}{4.1}
  \manuallabel{sec:multi_view}{4.2}
  \manuallabel{fig:learnedUV}{4}
  \manuallabel{tbl:more_qualitative_results}{4}
  \manuallabel{sec:visible_reconstruction}{5.1}
  \manuallabel{tbl:visible_recon}{1}

  \newcommand{\refpaper}[1]{in the paper}
\fi

\subsection{Overview}
In this supplementary, we provide additional details about our training (Sec.~\ref{sec:training}) and inference setups (Sec.~\ref{sec:inference}), and details of our evaluation metrics (Sec.~\ref{sec:metrics}). We provide an extended qualitative comparison of our method to the Image2Surf baseline (Sec.~\ref{sec:image2surf_comp}), ablations (Sec.~\ref{sec:ablation}) and for visible surface generation on real-world data (Sec.~\ref{sec:real-world}). We show additional qualitative results for hidden surface generation (Sec.~\ref{sec:hidden}) and also provide more visual results for Pix2Surf (Sec.~\ref{sec:more_results}) and more qualitative comparison to Pixel2Mesh++\cite{Wen:2019:pixel2mesh++} and AtlasNet\cite{Groueix:2018} (Sec.~\ref{sec:baseline_comparisons}).


\subsection{Training Details}
\label{sec:training}

For the \textbf{Single-View} case, we train our network in two phases. In the first phase, we train the NOCS-UV branch with a learning rate of \num{1e-4}, using the NOCS Map and the object mask as supervision. In the second phase, we add the remaining SP branch and train end-to-end until convergence, with a learning rate of \num{1e-4} for cars and \num{3e-5} for planes and chairs, and using the losses described in Sec.~\ref{sec:single_view}~\refpaper{}.

For the \textbf{Multi-View} case, we have found that pre-training with the single-view architecture, before switching to the full multi-view architecture results in better initialization. For this purpose, we start by passing the feature $z_m$, directly to the SP branch without max-pooling multiple views. After pre-training, we switch to the multi-view architecture as described in Sec.~\ref{sec:multi_view}~\refpaper{}, by max-pooling the $z_m$ features of all views, and concatenating both this max-pooled multi-view feature, and the single-view feature $z_m$ for the current view as input to the MLP. \jiahui{To better fuse multi-view information for learned chart prediction, the feature map in the middle of CNN encoder and decoder also follows above fusion operation.}
We randomly pick $5$ views as input during multi-view training. For our multi-view consistency loss, we need to identify corresponding pixels in different views. We sample pixels in each view as in the single-view case and find corresponding pixels based on their distance in NOCS coordinates. Two pixels are in correspondence if their NOCS distance is less than \num{1e-3}.

We separately train on each object category of our dataset.

\subsection{Inference Details}
\label{sec:inference}
\vspace{-0.3\baselineskip}
One significant advantage of our explicit \textbf{continuous} parametric surface prediction is that we can sample the results at any resolution (e.g. points or vertices). We generate our final predictions at a regular grid of samples in the unwrapped uv chart, obtaining a 3D location for each sample (obtained from the SP-Branch). Since we have exact correspondence to pixels of the input image, each sample also has a color value (or interpolated color value in super-resolution). Samples corresponding to background pixels are masked out. To create a mesh, we can connect neighboring foreground samples with edges.
All visual results of our method in the paper are generated using this approach. We provide more details.



\paragraph{Identifying foreground regions in the unwrapped chart.}
Unlike AtlasNet, the shape and topology of the unwrapped surface in our chart is learned by the NOCS-UV branch, which gives the reconstructed surface more flexibility to represent arbitrary shapes and topologies. To identify foreground regions in the uv space of the unwrapped chart, we map the the learned image-space foreground mask to uv space. 
Directly unwrap the mask by learned-uv map (two channel output from NOCS-UV branch) results in pixel cloud with holes in uv space. To solve this issue, we up-sample the image-space mask and learned-uv map from its original resolution of $240\times320$ by a factor of $4$ using linear interpolation before mapping mask to uv space. To avoid interpolating across $C^0$ discontinuities of the surface, we only interpolate neighboring pixels that are mapped to similar uv locations (i.e., the gradient of their uv coordinates is below a threshold). We then map the up-sampled mask to uv space (resolution of $128\times128$) by the up-sampled learned-uv map. Finally we up-sample the mask in uv space to the desired resolution (in paper we use $512\times512$).

In uv space, we additionally post-process the unwrapped foreground mask by closing small holes using morphological operations. Finally, we remove outliers using the predicted 3D locations (quarried from SP-Branch) of each mask sample. A sample of the foreground mask is classified as outlier if the distance in 3D space to its nearest neighbor is larger than a threshold $t$. In practice, we use $t=0.03$ for chairs and $t=0.02$ for cars and airplanes. \jiahui{Similar outlier removing operation is also applied to image-space mask before identifying foreground regions.}



\paragraph{Texturing the unwrapped chart.}
Similar to the mask, directly unwrapping the image-space color values to the uv space results in a sparse set of irregular color samples in uv space. 
We can interpolate these samples to obtain the color value at any point in uv space by interpolating the $k$ nearest neighbors (we use $k=4$ for our results).

\subsection{Evaluation Metrics}
\label{sec:metrics}
We now define the evaluation metrics used in the paper.

\textbf{A common surface representation}: Before evaluating our metrics, we convert the results of all methods to a common format to avoid biasing our results due to different surface representations. We convert all output representations to the NOCS-Map format defined in X-NOCS~\cite{Srinath:2019} using the ground truth camera model. The NOCS map $\mathcal{P}$ samples the reconstructed surface from a single viewpoint, giving a point cloud where each sample has a 2D pixel coordinate $p$ and a 3D location $x$. The 3D location is defined in a canonical coordinate frame that is shared across views and across instances of the same shape category. For multi-view reconstructions, we create one NOCS-Map for each viewpoint, compute the metrics on each NOCS-Map, and average the results over all views. \jiahui{As AtlasNet~\cite{Groueix:2018} ground truth is not in the same ShapeNet version as ShapeNet-Plain~\cite{Srinath:2019}, we first scale the AtlasNet results to have the same bounding box diagonal as the ground truth 2-intersection X-NOCS maps point cloud, and then align the lower left corner of the bounding box.}


The \textbf{Reconstruction Error} is measured as the 2-Way-Chamfer-Distance between the ground truth NOCS-Map $\mathcal{P}_{1}$ and predicted NOCS-Map $\mathcal{P}_{2}$:
$$
E_{\text{rec}} = \frac{1}{|\mathcal{P}_1|}\sum_{x_i \in \mathcal{P}_1}{\min_{y_j\in \mathcal{P}_2}\|x_i-y_j\|_2^2}+\frac{1}{|\mathcal{P}_2|}\sum_{y_j \in \mathcal{P}_2}{\min_{x_i\in \mathcal{P}_1}\|x_i-y_j\|_2^2}.
$$
The reconstruction error for hidden surfaces in Table 2 of paper is computed in the same way, but using NOCS-Maps of the hidden surfaces.

The \textbf{Correspondence Error} is measured as the squared distance between the predicted 3D location $x_i$ and the ground truth location $y_i$ of the same pixel:
$$
E_{\text{corr}} = \frac{1}{|\mathcal{M}|}\sum_{p_i \in \mathcal{M}}{\| x_i - y_i \|_2^2}.
$$
We only evaluate pixels $p_i 
\in \mathcal{M}$ that are both in the predicted and ground truth foreground masks.

\textbf{Consistency Error} is based on the squared distance between the predicted 3D locations of corresponding pixels in different views.
For each pair of views $a$ and $b$, we identify corresponding pairs of pixels $(p^a_i, p^b_j)$ as pairs having a similar ground truth 3D location in NOCS: $\|y^a_i-y^b_j\|_2<\epsilon$.
In practice, we set $\epsilon=0.001$.
We then average the squared distance between the predicted 3D locations $x^a_i$ and $x^b_j$ of all corresponding pixel pairs $\mathcal{P}^2_{\text{corr}}$:
$$
E_{\text{cons}} = \frac{1}{|\mathcal{P}^2_{\text{corr}}|}
\sum_{(p^a_i, p^b_j) \in \mathcal{P}^2_{\text{corr}}}{\|x^a_i-x^b_j\|_2^2}.
$$

With the \textbf{\jiahui{Dis}continuity Score},
we take a statistical approach to measure the \jiahui{correctness of the surface connectivity.} \jiahui{While the continuity of implicit or parametric surface is a property induced by representation and method design, we need to make sure the continuity is correct, i.e. no  over-smooth results. 
} We compute statistics of the $C^0$ discontinuities in the predicted surface, and measure the similarity to the same statistics computed on the ground truth surface. The statistics are based on the 3D distance $\| x_i - x_j \|_2$ of neighboring foreground pixels $p_i$ and $p_j$. Pixels with a large difference are likely to lie on the border of a $C^0$ discontinuity of the predicted surface. We compute a histogram $h$ of this 3D distance over all neighboring pixels:
$$
h_i = |\{(p_i,p_j) \in \mathcal{P}^2_{\text{neighbors}}\ |\ t_i \le \| x_i - x_j \|_2 < t_{i+1}\}|,
$$
where $t_i$ are the boundaries of the histogram bins and $\mathcal{P}^2_{\text{neighbors}}$ is the set of all neighboring pixel pairs. We use a $4$-connected neighborhood and choose $20$ bins with bin edges spaced uniformly in $[0.05, \sqrt{3}]$. We measure the similarity of two histograms as the correlation of their normalized bins:
$$
S_{\text{cont}} = \frac{1}{\sum_k h_k \sum_j h^{\text{gt}}_j} \sum_i h_i h^{\text{gt}}_i
$$
Note that unlike the other errors we use as evaluation metrics, this is defined as a score, where higher values imply more accurate \jiahui{dis}continuity of the reconstructed surface.

\begin{figure*}[!b]
    \centering
    \includegraphics[width=0.96\columnwidth]{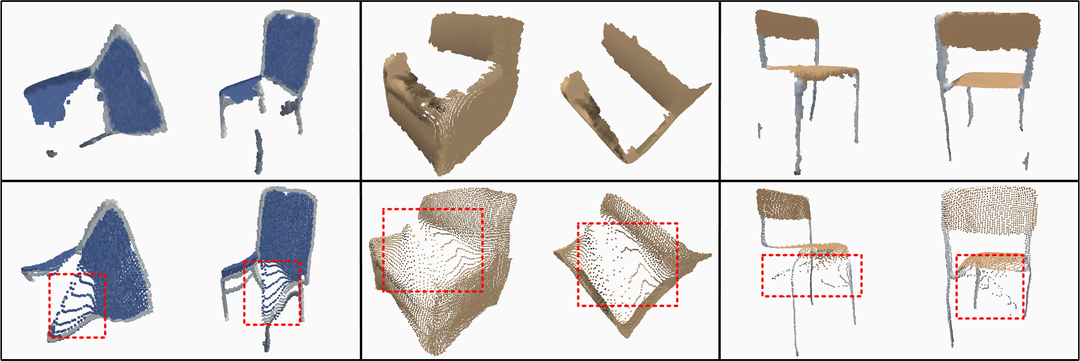}
    \caption{Qualitative Comparison to Image2Surf. The first row are the results of Pix2Surf and the second row are for Image2Surf. Each instance is viewed from 2 different viewpoints. Image2Surf wrongly connects disjoint parts and results in strong distortions, which are solved by Pix2Surf's learned chart.}
    \label{fig:image2surf_comp}
\end{figure*}

\subsection{Qualitative Comparison to Image2Surf}
\label{sec:image2surf_comp}
We show more qualitative comparisons between our baseline Image2Surf and Pix2Surf in paper Sec. 5.1. Image2Surf has a fatal problem to make ``cut" around the occlusion boundary (i.e., wrong $C^0$ discontinuities), which is reflected both in the red rectangle in Fig.~\ref{fig:image2surf_comp} and \jiahui{dis}continuity score in Table 1 in paper.

\subsection{Ablations}
\label{sec:ablation}
We provide an analysis and justification of several key design choices.
First, we analyze the importance of using a learned $UV$ chart instead of a fixed chart like Image2Surf.
As seen in Sec.~\ref{sec:visible_reconstruction} and Table~\ref{tbl:visible_recon}~\refpaper{}, Pix2Surf outperforms Image2Surf on all categories for reconstruction error.
Second, we analyze the utility of multi-view feature pooling and consistency loss.
As seen in Table~\ref{tbl:visible_recon} (rows 4--7), these two features significantly improve performance.
We also justify the use of intermediate NOCS map regression by the NOCS-UV branch, and the need for the UV amplifier (Table~\ref{tbl:single_ablation}).
We do so by examining networks without these two components.
For the NOCS map ablation, we train the network from scratch without pretraining the NOCS-UV branch, and for the UV amplifier ablation, we directly input the learned UV coordinates to the SP branch and increase the dimension of a latent image code to 256.
When conducting the experiments on the chair category, the results (Table~\ref{tbl:single_ablation}) show that these components help learn better reconstructions.

\begin{table}
\centering
\caption{We experimentally verify the usefulness of NOCS map regression and the UV amplifier.
NOCS map regression provides intermediate supervision while the UV amplifier balances information.
Here we report average reconstruction error computed on the \emph{visible} part (equal training epochs for all methods).}
\label{tbl:single_ablation}
{\small
\setlength{\tabcolsep}{0.2em}
\renewcommand{\arraystretch}{0.9}
\newcolumntype{Y}{>{\centering\arraybackslash}X}
\begin{tabularx}{\linewidth}{Y|Y|Y|Y}
  \toprule
     & No UV Amp. & No NOCS & Pix2Surf \\
  \midrule
    Chair & 10.37 & 3.64 & \textbf{2.61} \\
  \bottomrule
\end{tabularx}
}
\end{table}

\subsection{Qualitative Results on Real-World Data}
\label{sec:real-world}
\begin{figure*}[!t]
  \centering
  \renewcommand{\arraystretch}{2}
  \begin{tabular}{l|r}
    \includegraphics[width=0.49\textwidth]{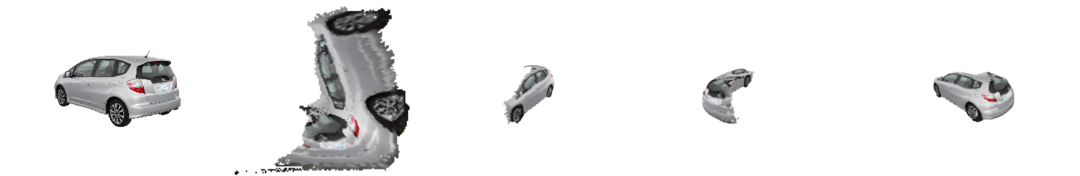} &
    \includegraphics[width=0.49\textwidth]{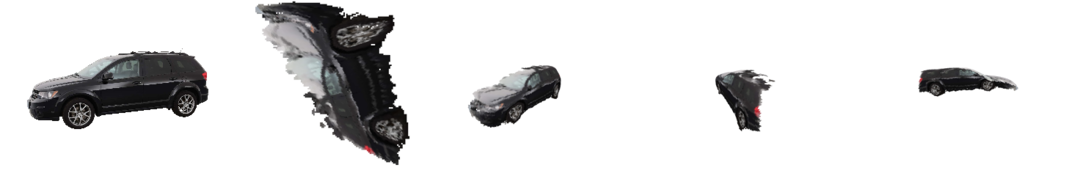} \\
    \includegraphics[width=0.49\textwidth]{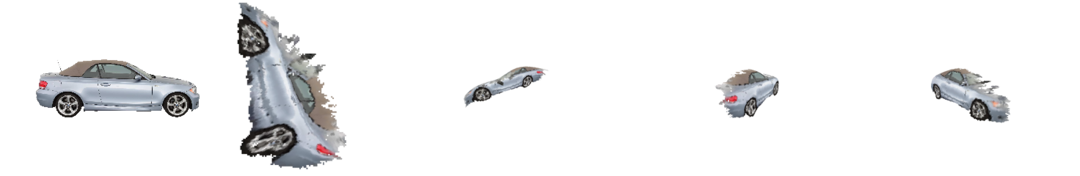} &
    \includegraphics[width=0.49\textwidth]{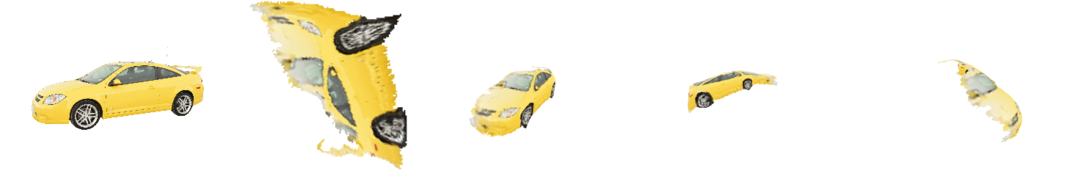} \\
    \midrule
    \includegraphics[width=0.49\textwidth]{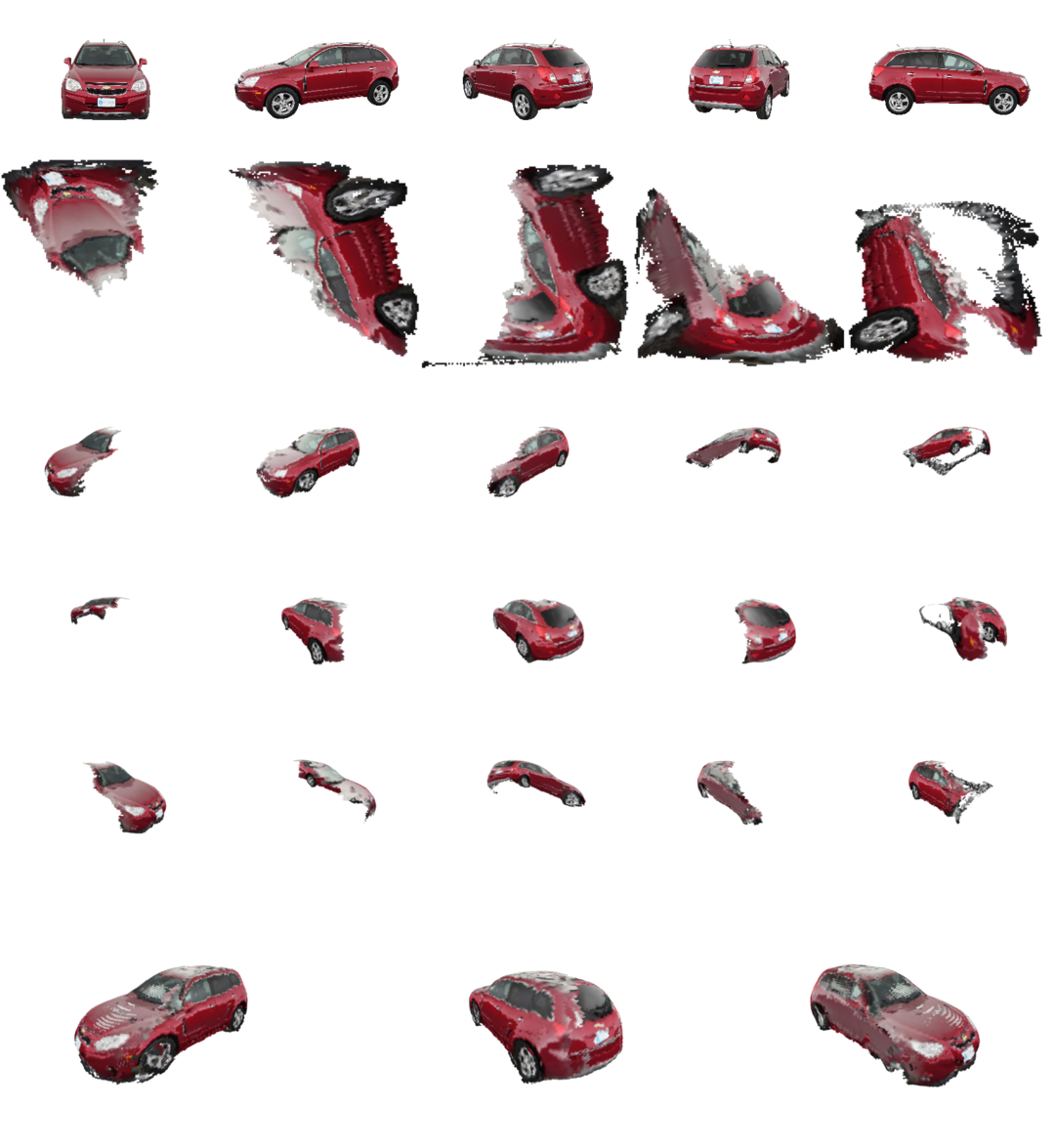} &
    \includegraphics[width=0.49\textwidth]{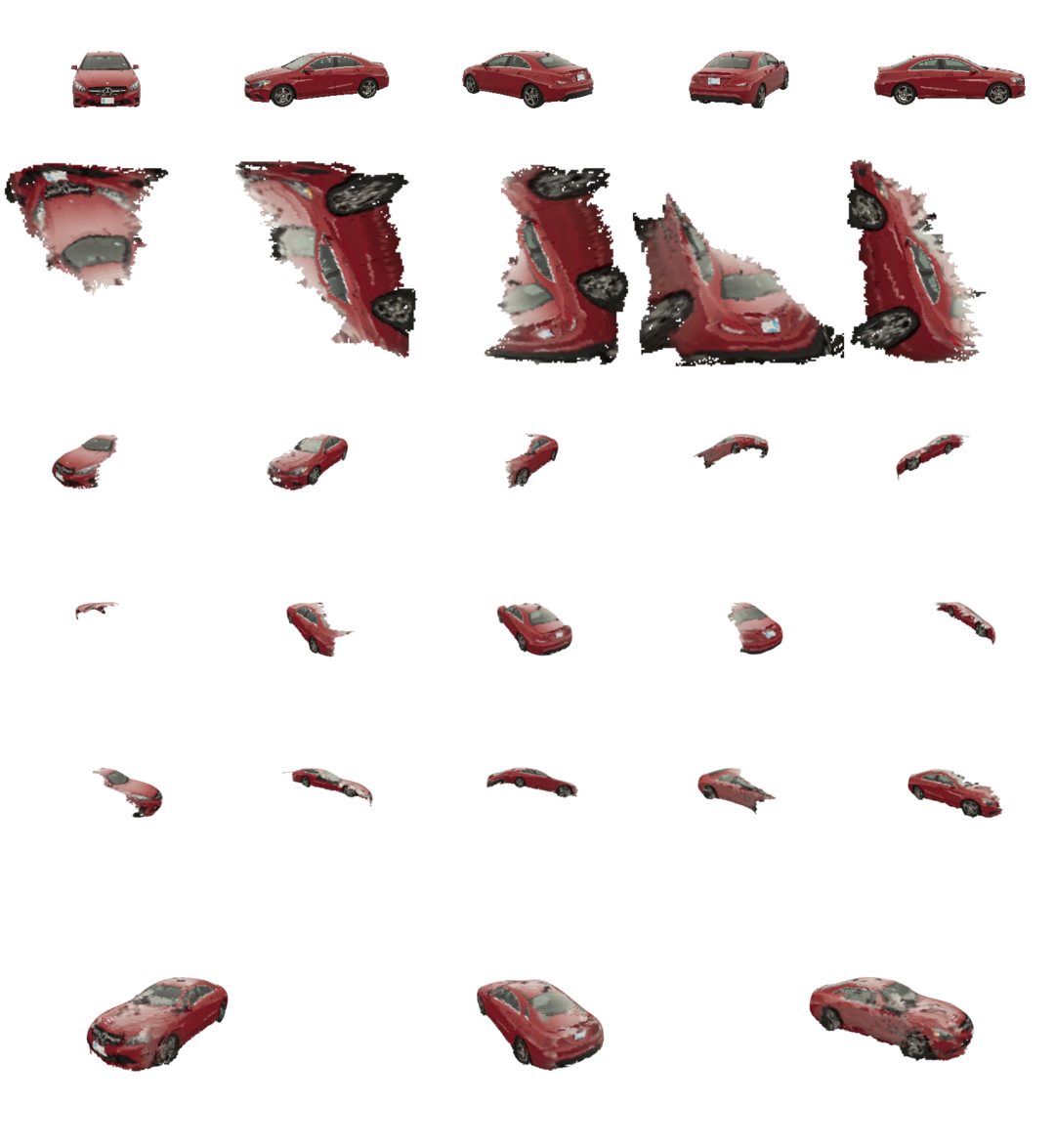} \\
  \end{tabular}
  \caption{Real world image generalization. The top part is single-view visualization: input image, unwrapped chart with texture and 3 viewpoints of the reconstruction for each instance. 
  The bottom part is multi-view aggregation visualization. For every instance, each row is: input images, unwrapped charts with texture, 3 viewpoints for each view's result separately and finally multi view aggregation.}
  \label{fig:real_world}
\end{figure*}

We show more results for generalization to real world data mentioned in paper. In Fig.~\ref{fig:real_world}, we show single- and multi-view results for Pix2Surf that is trained on ShapeNet COCO and inference on real world car. Note that the texture in each view separately is better than the multi-view aggregation. This is caused by the different light condition from different viewpoints. As our main concern in this paper is not to fuse the texture from multiple views, we leave the improvement of the texture to future works.

\input{content/supp/text/11_more_results}

\clearpage
\newpage
\subsection{Qualitative Results for Hidden Surface Generation}
\label{sec:hidden}
The following table provides more visual results of Pix2Surf Two-Intersection version (Sec 5.2 in paper), and comparison with X-NOCS~\cite{Srinath:2019}. Pix2Surf can easily be extended to capture the invisible surface and is more accurate and smooth than X-NOCS.

\newcommand{\AllTwoIntersectionImages}{\input{content/supp/figure/two_intersection/list.tex}}
\graphicspath{{content/supp/figure/two_intersection/}}

\makeatletter
\def\Image#1{%
  \multicolumn{\LT@cols}{l}{\includegraphics[width=\textwidth]{#1}}\\
}
\makeatother

\setlength{\tabcolsep}{0em}
\def\arraystretch{0.0}
\newcolumntype{Y}{>{\centering\arraybackslash}m{0.1111\textwidth}}
{\scriptsize
\begin{longtable}{YYY|YYY|YYY}
  \toprule
  \multicolumn{3}{c|}{View 1} &
  \multicolumn{3}{c|}{View 2} &
  \multicolumn{3}{c}{View 3} \\
  \midrule
  \makecell{Pix2Surf\\(sv)} & \makecell{X-NOCS\\(sv)} & \makecell{Ground\\Truth} &
  \makecell{Pix2Surf\\(sv)} & \makecell{X-NOCS\\(sv)} & \makecell{Ground\\Truth} &
  \makecell{Pix2Surf\\(sv)} & \makecell{X-NOCS\\(sv)} & \makecell{Ground\\Truth} \\
  \midrule
  \endhead

  \bottomrule
  \endfoot

  \input{content/supp/figure/two_intersection/list.tex}
\end{longtable}
}

\input{content/supp/text/12_comparisons}

%% file: content/supp/text/11_more_results.tex

\clearpage
\newpage
\subsection{More Results}
\label{sec:more_results}
\begin{figure*}[h]
  \centering
  \renewcommand{\arraystretch}{2}
  \begin{tabular}{l|r}
  
    \includegraphics[width=0.49\textwidth]{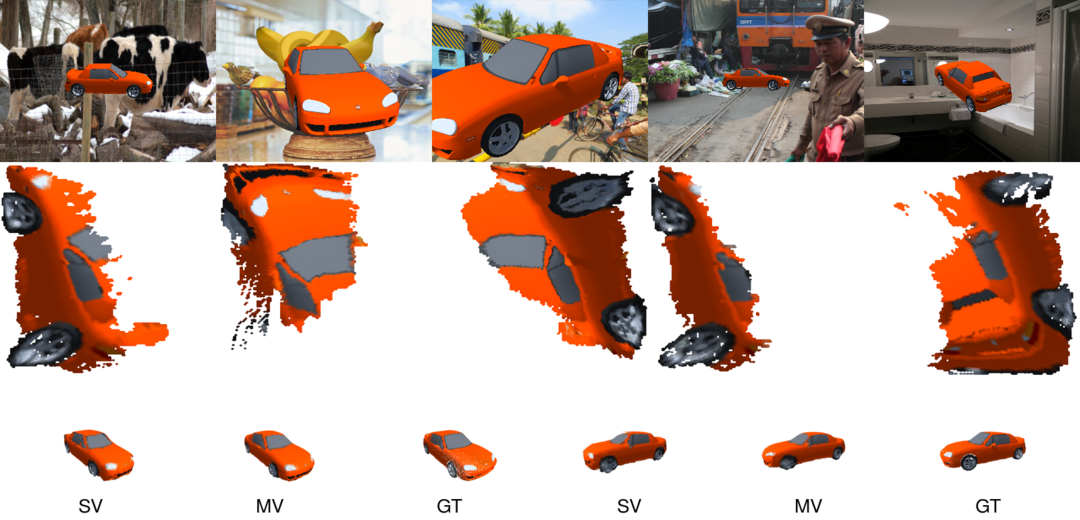} &
    \includegraphics[width=0.49\textwidth]{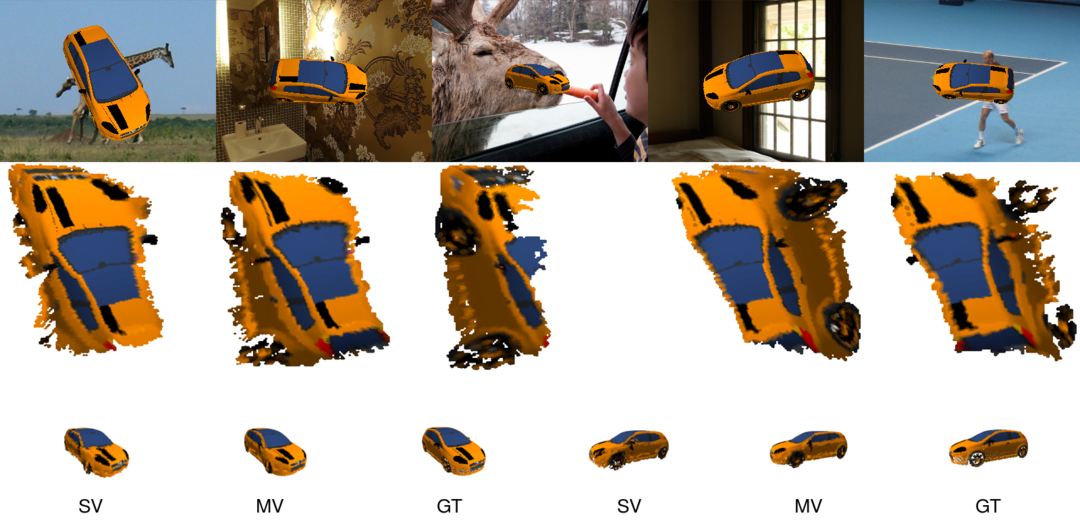} \\
    \includegraphics[width=0.49\textwidth]{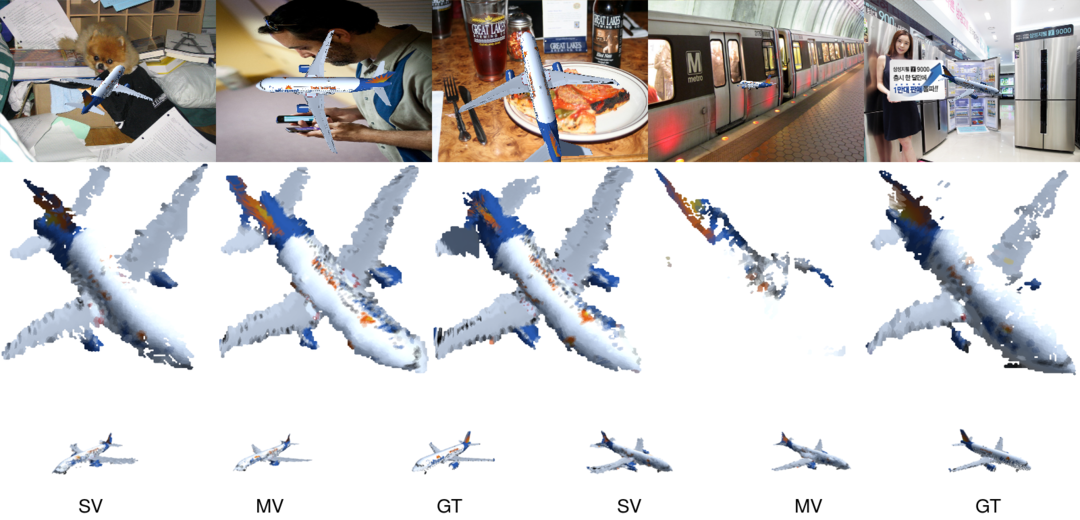} &
    \includegraphics[width=0.49\textwidth]{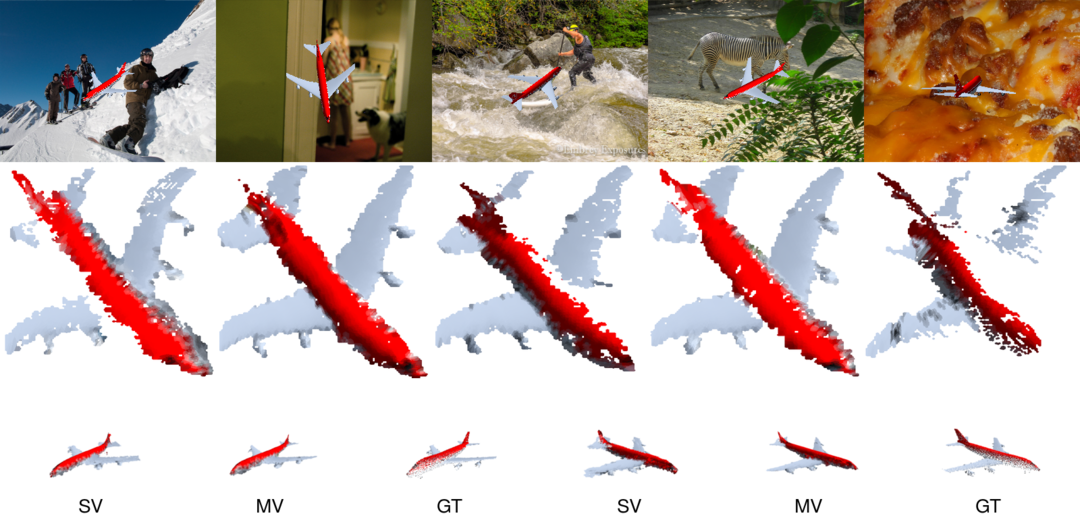} \\
    \includegraphics[width=0.49\textwidth]{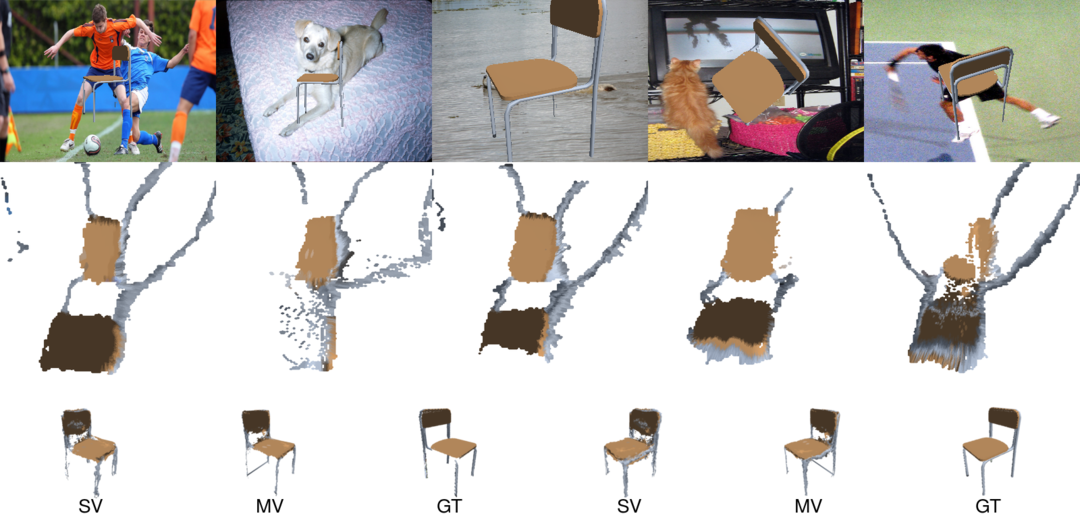} &
    \includegraphics[width=0.49\textwidth]{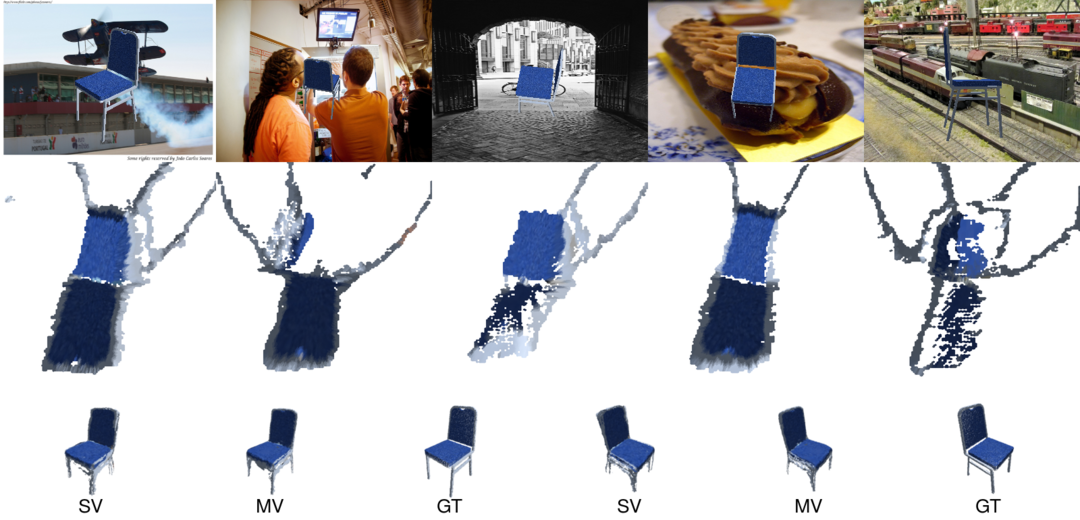} \\
  \end{tabular}
  \caption{Single-view and multi-view Pix2Surf reconstruction results. The results for each object are presented in three rows. The first row shows five input views. Note that we do not have camera parameters for any of these views. The second row shows the per-view UV space that is generated by the multi-view variant of Pix2Surf. The UV space is not directly constrained by any loss; the flattening of the objects that we can observe and the large degree of consistency between different views is an emergent property of our network. In the third row, we show, from left to right, (a) the reconstructed 3D surface obtained by merging Pix2Surf single-view reconstructions (SV), (b) the Pix2Surf multi-view reconstruction (MV), and (c) the ground truth reconstruction (GT). The last three columns show the same results from a different viewpoint. Note the reduction in the number of gaps and surface discontinuities when comparing the multi-view to the single-view results.}
  \label{fig:more_results}
\end{figure*}

Figure~\ref{fig:more_results} shows more results of Pix2Surf including the learned UV map (as shown in Figure~\ref{fig:learnedUV}) and reconstruction outputs of both single-view and multi-view architectures. See the caption for the details.

%% file: content/supp/figure/two_intersection/list.tex
\Image{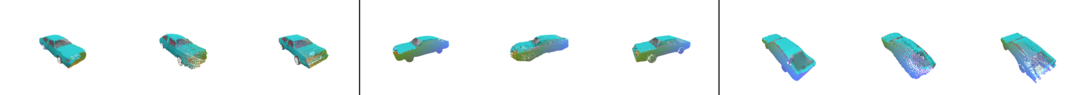}
\Image{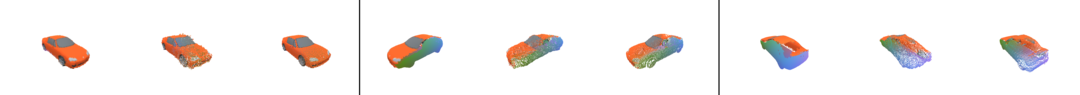}
\Image{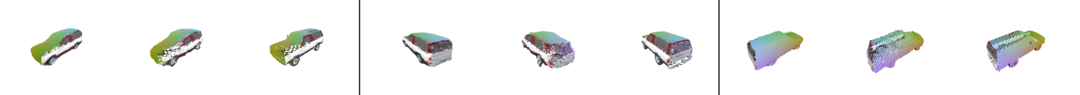}
\Image{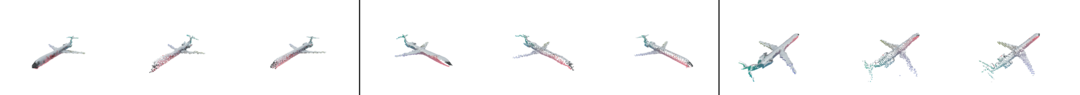}
\Image{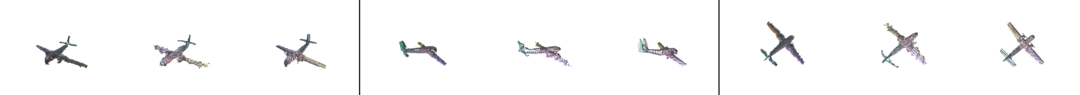}
\Image{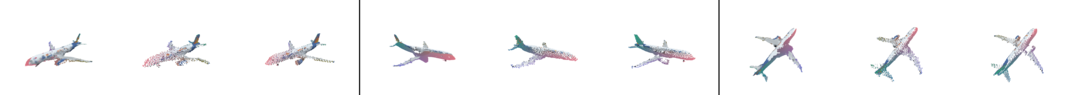}
\Image{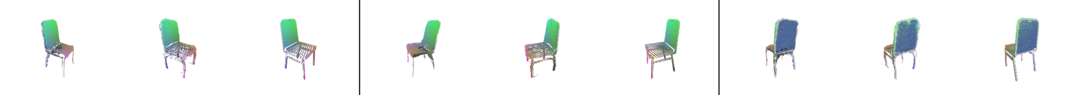}
\Image{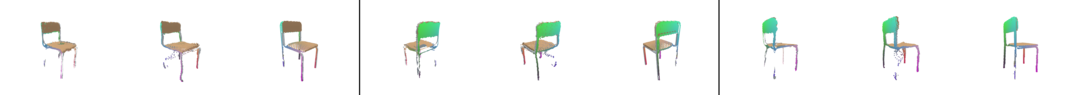}
\Image{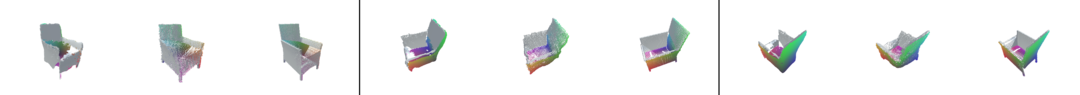}

%% file: content/supp/text/12_comparisons.tex

\clearpage
\newpage
\subsection{Qualitative Comparisons}
\label{sec:baseline_comparisons}

The following table demonstrates qualitative comparisons among our Pix2Surf (both single-view and multi-view architectures), AtlasNet\cite{Groueix:2018}, and Pixel2Mesh++~\cite{Wen:2019:pixel2mesh++}. The colors in AtlasNet results show different output patches.

\newcommand{\AllComparisonImages}{\input{content/supp/figure/comparisons/comparisons.tex}}
\graphicspath{{content/supp/figure/comparisons/}}

\makeatletter
\def\Image#1{%
  \multicolumn{\LT@cols}{l}{\includegraphics[width=\textwidth]{#1}}\\
}
\makeatother

\setlength{\tabcolsep}{0em}
\def\arraystretch{0.0}
\newcolumntype{Y}{>{\centering\arraybackslash}m{0.1249\textwidth}}
{\scriptsize
\begin{longtable}{YYY|YYY|YY}
  \toprule
  \multicolumn{3}{c|}{View 1} &
  \multicolumn{3}{c|}{View 2} &
  \multicolumn{2}{c}{View 1} \\
  \midrule
  \makecell{Single\\View} & \makecell{Multi-\\View} & \makecell{Ground\\Truth} &
  \makecell{Single\\View} & \makecell{Multi-\\View} & \makecell{Ground\\Truth} &
  \makecell{Atlas\\Net} & \makecell{Pixel2\\Mesh++} \\
  \midrule
  \endhead

  \bottomrule
  \endfoot

  \input{content/supp/figure/comparisons/comparisons.tex}
\end{longtable}
}

%% file: content/supp/figure/comparisons/comparisons.tex
\Image{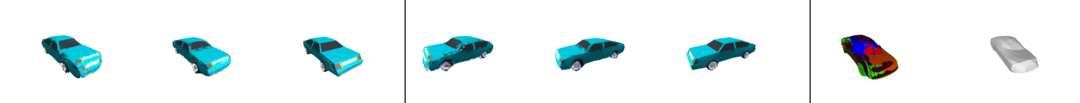}
\Image{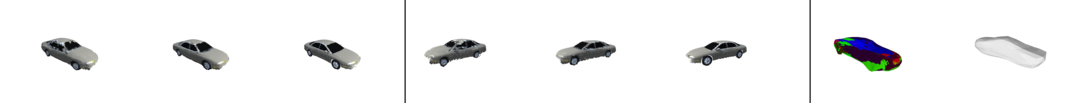}
\Image{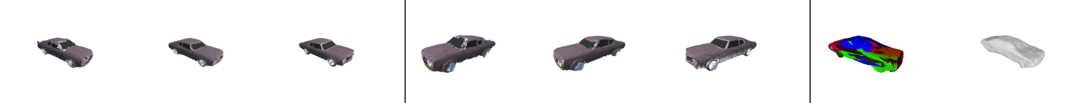}
\Image{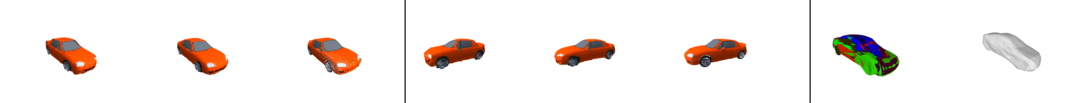}
\Image{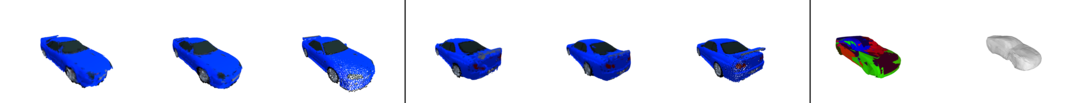}
\Image{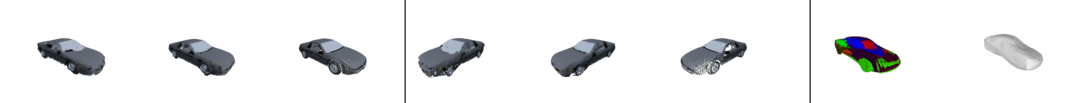}
\Image{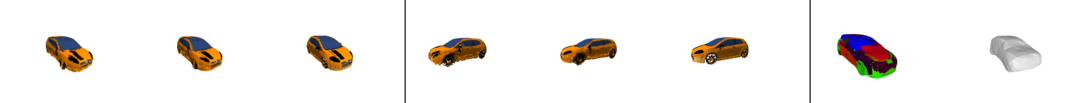}
\Image{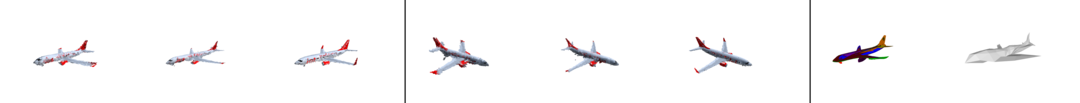}
\Image{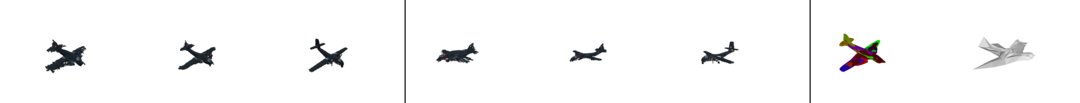}
\Image{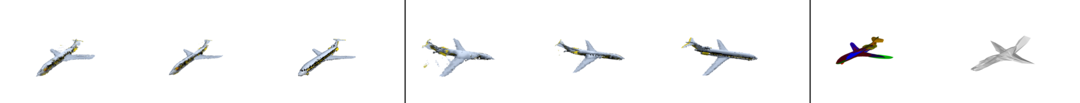}
\Image{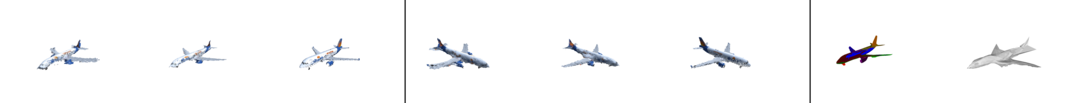}
\Image{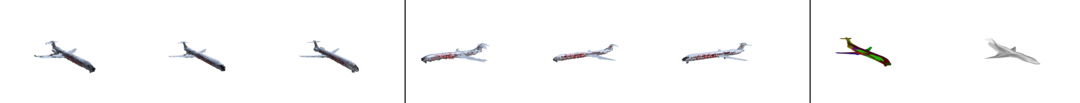}
\Image{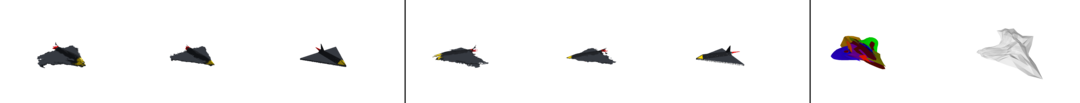}
\Image{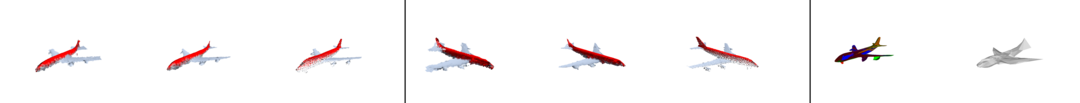}
\Image{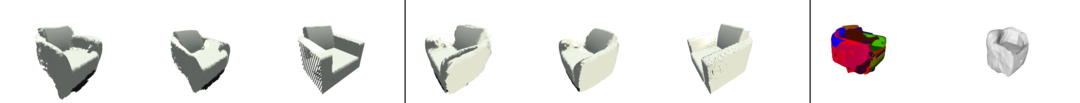}
\Image{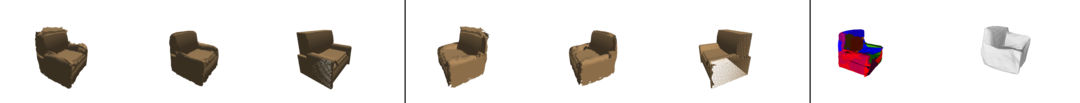}
\Image{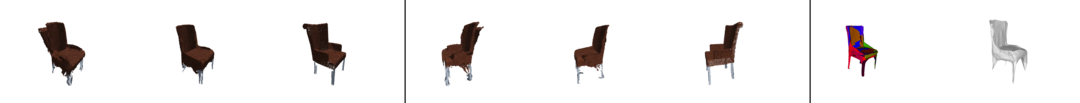}
\Image{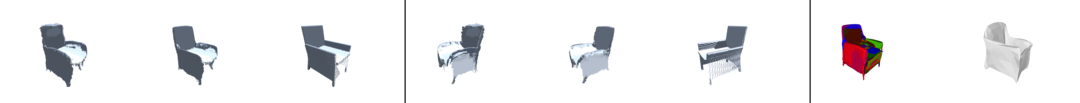}
\Image{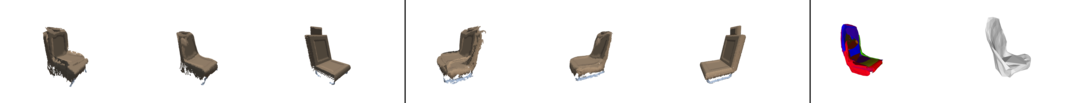}
\Image{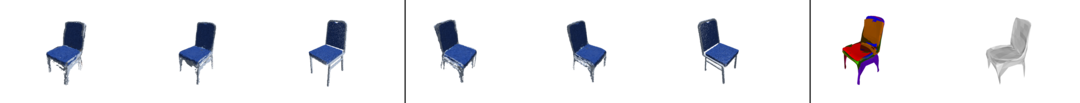}
\Image{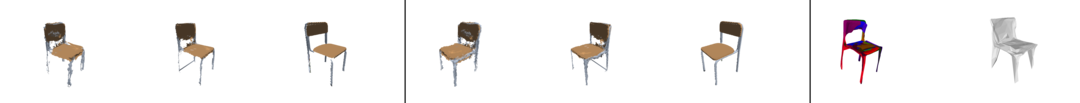}